\documentclass{article}

\usepackage[preprint]{corl_2024} 

\usepackage{amsmath}
\usepackage{amssymb}
\usepackage{booktabs}
\usepackage{iitem}
\usepackage{multirow}
\usepackage{pifont}
\usepackage{graphicx}
\usepackage{siunitx}
\usepackage{wrapfig}
\usepackage{enumitem}
\usepackage{calc}
\usepackage{titlesec}
\usepackage{threeparttable}

\usepackage{tikz}
\usepackage{pgfplots}
\usetikzlibrary{positioning}
\usetikzlibrary{calc}
\usetikzlibrary{spy}
\pgfplotsset{compat=newest}

\definecolor{my_blue}{rgb}{0, 0.4470, 0.7410}
\definecolor{my_yellow}{rgb}{0.9290, 0.6940, 0.1250}
\definecolor{my_purple}{rgb}{0.4940, 0.1840, 0.5560}
\definecolor{my_green}{rgb}{0.4660, 0.6740, 0.1880}
\definecolor{my_red}{rgb}{0.6350, 0.0780, 0.1840}
\definecolor{my_black}{rgb}{0.25, 0.25, 0.25}
\definecolor{my_turquoise}{rgb}{0.000, 0.710, 0.694}
\definecolor{my_pink}{rgb}{0.910, 0.263, 0.576}

\newcommand\figref{Figure~\ref}
\newcommand\tabref{Table~\ref}

\newcommand{\robotname}{Berkeley Humanoid}

\title{\robotname: A Research Platform for \\ Learning-based Control}

\author{
  Qiayuan Liao\\
  \texttt{qiayuanl@berkeley.edu}
  \And
  Bike Zhang\\
  \texttt{bikezhang@berkeley.edu}
  \And
  Xuanyu Huang\\
  \texttt{xuanyuhuang2001@gmail.com}
  \And
  Xiaoyu Huang\\
  \texttt{x.h@berkeley.edu}
  \And
  Zhongyu Li\\
  \texttt{zhongyu\_li@berkeley.edu}
  \And
  Koushil Sreenath\\
  \texttt{koushils@berkeley.edu}
}

\usepackage{pgfplots}
\usepgfplotslibrary{groupplots}
\usetikzlibrary{spy}

\pgfplotsset{
    legend cell align={left},
    legend style={font=\scriptsize},
    label style={font=\footnotesize, inner sep=0pt},
    legend cell align={left},
    tick label style={font=\footnotesize}
}

\newcommand{\plottracking}{
\begin{tikzpicture}[spy using outlines={rectangle, magnification=3, connect spies}]
\begin{groupplot}[
    group style={
        group name=plots,
        group size=1 by 3,
        vertical sep=5pt},
    width=0.75\linewidth,
    height=0.3\linewidth,
    xmin=3, xmax=50,
    ylabel =\empty,
    legend cell align={left},
    legend style={font=\scriptsize},
    every axis plot/.append style={
        line width=0.8pt}
]
\nextgroupplot[ylabel = {$\mathbf{v}_x$ [m/s]}, xticklabel=\noexpand\empty, legend pos=outer north east]
\addplot[my_blue] table[col sep=comma, x expr=\thisrow{__time}, y expr=\thisrow{/cmd_vel/linear/x}]{data/random_mat_cmd_vel.csv};
\addplot[my_yellow,opacity=0.3] table[col sep=comma, x expr=\thisrow{__time}, y expr=\thisrow{/odom/twist/twist/linear/x}]{data/random_mat_odom_5_downsampled.csv};
\addplot[my_yellow] table[col sep=comma, x expr=\thisrow{__time}, y expr=\thisrow{/odom/twist/twist/linear/x_smoothed}]{data/random_mat_odom_5_downsampled.csv};
\addplot[my_purple,opacity=0.3] table[col sep=comma, x expr=\thisrow{__time}+2.392580986022949, y expr=\thisrow{/isaac_lab/twist/twist/linear/x}]{data/lin_vel_log_0.5_5_downsampled.csv};
\addplot[my_purple] table[col sep=comma, x expr=\thisrow{__time}+2.392580986022949, y expr=\thisrow{/isaac_lab/twist/twist/linear/x_smoothed}]{data/lin_vel_log_0.5_5_downsampled.csv};

\node[align=right, anchor=east, font=\footnotesize] at (axis cs:50, -0.6) {Error: {\color{my_purple} 0.051 m/s}\\{\color{my_yellow} 0.058 m/s}};

\legend{Command, Real (raw), Real (smoothed), Isaac Lab(raw), Isaac Lab (smoothed)}

\nextgroupplot[ylabel = {$\mathbf{v}_y$ [m/s]}, xlabel={Time [s]}]
\addplot[my_blue] table[col sep=comma, x expr=\thisrow{__time}, y expr=\thisrow{/cmd_vel/linear/y}]{data/random_mat_cmd_vel.csv};
\addplot[my_yellow,opacity=0.3] table[col sep=comma, x expr=\thisrow{__time}, y expr=\thisrow{/odom/twist/twist/linear/y}]{data/random_mat_odom_5_downsampled.csv};
\addplot[my_yellow] table[col sep=comma, x expr=\thisrow{__time}, y expr=\thisrow{/odom/twist/twist/linear/y_smoothed}]{data/random_mat_odom_5_downsampled.csv};
\addplot[my_purple,opacity=0.3] table[col sep=comma, x expr=\thisrow{__time}+2.392580986022949, y expr=\thisrow{/isaac_lab/twist/twist/linear/y}]{data/lin_vel_log_0.5_5_downsampled.csv};
\addplot[my_purple] table[col sep=comma, x expr=\thisrow{__time}+2.392580986022949, y expr=\thisrow{/isaac_lab/twist/twist/linear/y_smoothed}]{data/lin_vel_log_0.5_5_downsampled.csv};

\node[align=right, anchor=east, font=\footnotesize] at (axis cs:50, -0.6) {Error: {\color{my_purple} 0.086 m/s}\\{\color{my_yellow} 0.116 m/s}};

\coordinate (spypoint) at (axis cs:41, 0.48);
\coordinate (magnifyglass) at (axis cs:59, 0.0);
\spy[my_green,width=0.22\linewidth, height=0.15\linewidth, every spy on node/.append style={thick}] on (spypoint) in node [fill=white] at (magnifyglass);

\end{groupplot}
\node[above=0.075\linewidth of magnifyglass, align=right,anchor=south,font=\footnotesize] {Similiar steady-state error};

\end{tikzpicture}
}

\newcommand{\plotgps}{
\begin{tikzpicture}
    \newlength{\height}
    \setlength{\height}{0.55\linewidth}
    \begin{axis}[
        name=plot,
        height=0.7\height,
        xlabel = {Time [s]},
        ylabel = {Elevation [m]}, 
        enlargelimits=0.05,
    ]
    \addplot[my_blue, line width=1.0pt] table[col sep=comma, x expr=\thisrow{time_seconds}, y expr=\thisrow{ele}]{data/track_gps_modified.csv};
    \end{axis}
    \node[left=30pt of plot] (map) {\includegraphics[height=\height, angle=90]{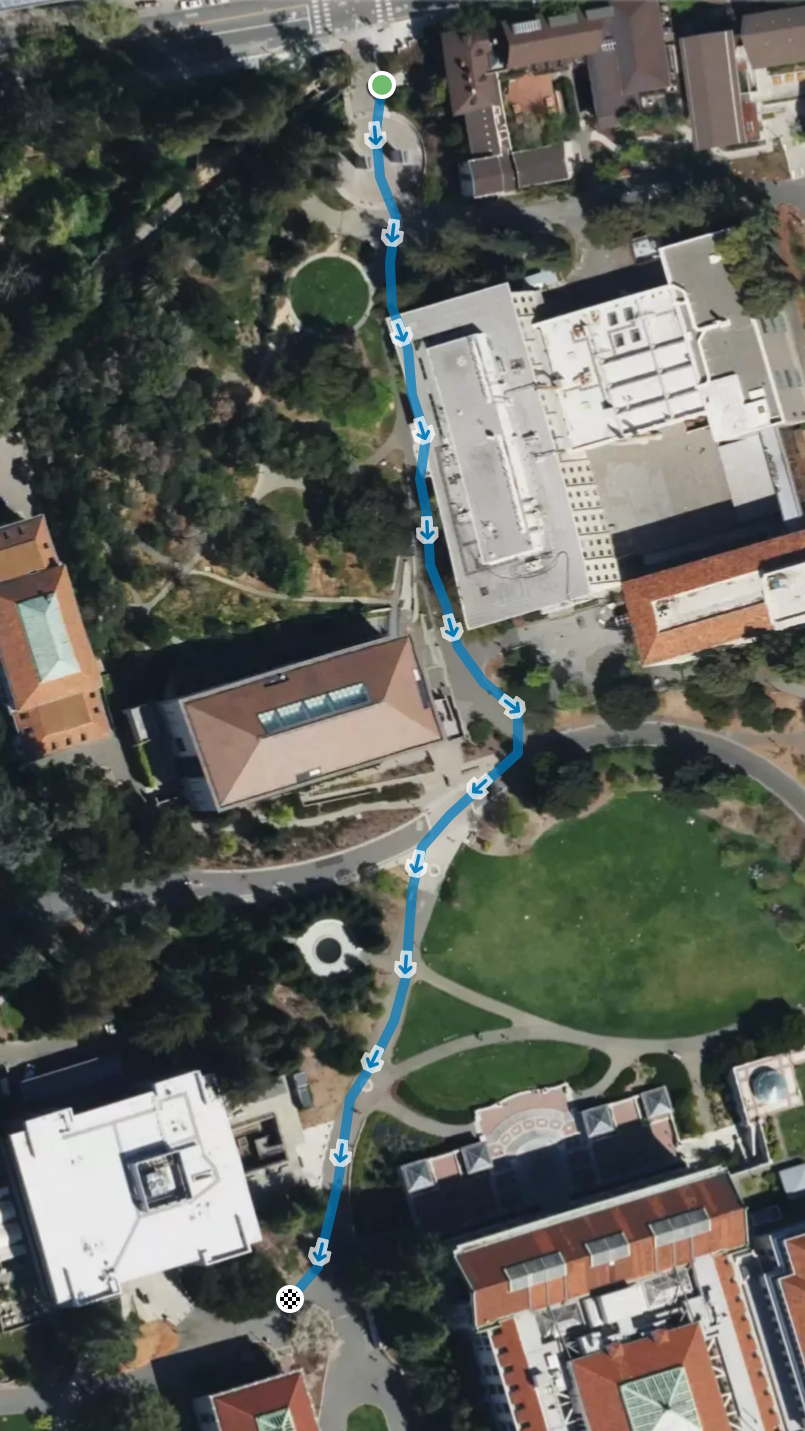}};
\end{tikzpicture}
}

\begin{document}

\maketitle

\begin{figure}[ht]
    \centering
    \includegraphics[width=1.0\linewidth]{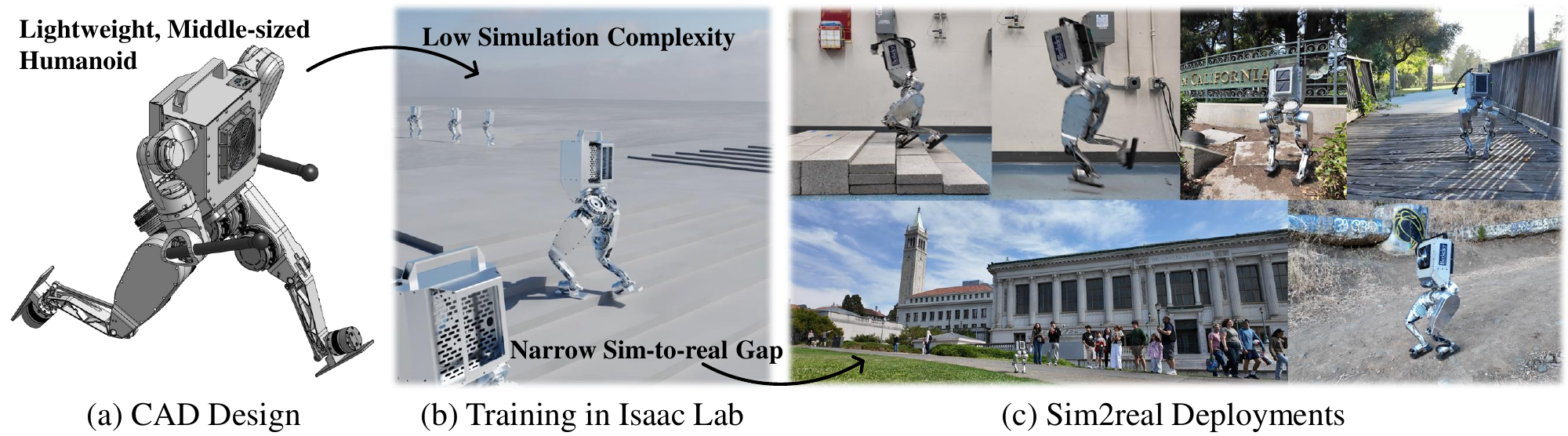}
    \caption{Design, training, and sim-to-real deployment of our custom-built humanoid with a learning-based controller.}
    \label{fig:teaser}
\end{figure}


\begin{abstract}
We introduce \robotname, a reliable and low-cost mid-scale humanoid research platform for learning-based control. Our lightweight, in-house-built robot is designed specifically for learning algorithms with low simulation complexity, anthropomorphic motion, and high reliability against falls. 
The robot's narrow sim-to-real gap enables agile and robust locomotion across various terrains in outdoor environments, achieved with a simple reinforcement learning controller using light domain randomization.
Furthermore, we demonstrate the robot traversing for hundreds of meters, walking on a steep unpaved trail, and hopping with single and double legs as a testimony to its high performance in dynamical walking. 
Capable of omnidirectional locomotion and withstanding large perturbations with a compact setup, our system aims for scalable, sim-to-real deployment of learning-based humanoid systems. Please check our \href{http://berkeley-humanoid.com/}{website} for more details.
\end{abstract}



\section{Introduction}
\label{sec:intro}
There is a strong need for mid-scale humanoid robots that are designed to enable fast deployment of learning-based policies, robust to falls and failures, and inexpensive, with the ability to perform highly dynamic motions.
Most current bipedal and humanoid robots \cite{englsberger2014overview, seiwald2021lola, tsagarakis2017walk, digit, unitreeh1} are larger, unsafe, and require a team of people to operate.
Experiments with shorter-legged robots \cite{ramos2018facilitating, katz2018low, chignoli2021humanoid, katz2019mini, liu2022design,grimminger2020open, ghansah2024dynamic} are easier as they do not require a gantry crane due to their lightweight nature, allowing them to be carried by one person. Falls typically do not damage the environment or the robot, making the setup more forgiving. These robots can be tested in cramped lab spaces, and moreover, creating rough terrain for experimental validation is relatively simple due to their limited ground clearance.
We believe there is a substantial demand for mechanically reliable, inexpensive, short-legged humanoid robots with custom high-torque density actuators that is designed for rapid learning policy iteration.

Mechanical design is more challenging for shorter-legged humanoid robots due to limited space for housing components such as motors, sensors, and wiring, necessitating the use of compact power-dense actuators that are often very expensive or not available off-the-shelf \cite{hutter2016anymal, katz2018low, hattori2020design}. Integrating all components in a compact volume without sacrificing performance or cost is difficult. 
Furthermore, mid-scale robots are more handy and often leveraged to push the limits of dynamic and agile tasks~\cite{katz2019mini}, requiring an even higher torque-to-weight ratio and greater impact reliability.

Control for mid-scale humanoids is more challenging due to their low center of gravity and heightened sensitivity to disturbances, which lead to instability. Their lower mass and inertia make these robots more agile but also more sensitive with even small forces producing large motions. The shorter legs result in a reduced stride length, often necessitating multiple steps to counteract perturbations. Additionally, these robots require higher frequency leg movements to adjust foot placement rapidly, demanding precise coordination and control. 
These characteristics mean that the actuation of the joints must be quick and accurate to support high-frequency motions, and the control policies need to be exceptionally precise and robust to match the short-time constants of the dynamics. 
Furthermore, learning-based algorithms predominant in humanoid control face substantial sim-to-real gaps, particularly in executing rapid and dynamic motions that are required for controlling these robots. Consequently, utilizing learning-based control for mid-scale humanoids presents additional challenges.

To address these problems, we propose to custom-build a mid-scale humanoid platform with a special emphasis on facilitating learning-based control.
To achieve accurate, robust, and agile control, we leverage a learning-based algorithm and focus on narrowing the sim-to-real gap with adequate hardware design.
Learning-based algorithms enable us to leverage cheaper, and noisier sensors to cut down costs. 
To optimize for simulation performance and accuracy while achieving high-performance actuation, we utilize custom modular actuators with integrated transmission, hollow shafts, and EtherCAT for communication. 
Our contributions are summarized as follows: 
(i) We present a reliable, low-cost, mid-scale humanoid research platform focusing on narrowing the sim-to-real gap with design considerations tailored for learning-based control. 
(ii) We demonstrate that our design choices facilitate us to be able to use a minimally composed control policy to perform dynamic and robust locomotion on complex terrain, notably the challenging task of walking on a steep, narrow, and unpaved trail.
(iii) The codebase for policy training with the recent Isaac Lab release will be open-sourced to support future humanoid research upon acceptance.

\section{Related Work}
\label{sec:citations}

\paragraph{Humanoid Design.}
As shown in \tabref{tab:compare_to_other_robots}, we categorize humanoid robots into three primary sizes: (a) full-scale, which corresponds to the size of an average adult, (b) mid-scale, comparable to the size of a child, and (c) miniature, which refers to tiny non-human-sized robots.
full-scale humanoid or biped research platforms typically have a large weight and use high gear ratio Harmonic Drive actuators \citep{englsberger2014overview, tsagarakis2017walk,seiwald2021lola}. These platforms are primarily capable of walking and performing simple arm manipulations. Some platforms utilize Cycloidic Drive Actuators for high-load joints, combined with spring and linkage designs \citep{digit, cassie}. 
This setup simplifies the design of reduced-order, step-to-step model-based controllers.
However, for more recent learning-based algorithms, these designs optimized for model-based control advertedly affect training and deployment.
In comparison, more lightweight platforms featuring Quasi-Direct-Drive (QDD) actuators and primarily dummy arms have been recently developed, capable of performing more dynamic tasks \citep{unitreeh1, zhu2023design}.
Besides full-scale humanoids, mid-scale or miniature humanoid research platforms have gained popularity over the recent years \citep{li2023dynamic, liu2022design, saloutos2023design, unitreeg1,wang2015hermes}. All of these platforms opt for QDD actuators and are designed for better dynamic performance, but most of them lack fully articulated legs.
On the other hand, new humanoid robots from some companies deviate from QDD: Tesla Optimus, for example, uses linear actuators and harmonic drives, some with load cells for force control, and features complex transmissions between joints and actuators \citep{khan2015bio}. Boston Dynamics' hydraulic Atlas \citep{atlas} excels in highly dynamic tasks, and the newly released electric Atlas~\citep{atlas} showcases simplified joint designs with a large range of motion. The robots from companies are well-designed and well-tested, but unfortunately, most of them are not available for researchers in labs or do not provide access to modify or improve the low-level system. 

\paragraph{Humanoid Control.}
Humanoid control is a challenging problem in the robotics field. Utilizing control approaches ranging from heuristic-based methods to model-based control, humanoids have been equipped with stable movement abilities \citep{raibert1986legged,kajita20013d,kuindersma2016optimization,ding2022orientation,khazoom2024tailoring}. Recently, learning-based approaches demonstrate promising capabilities for humanoid robots, ranging from locomotion 
\citep{siekmann2021sim,radosavovic2024real,singh2023learning,tang2023humanmimic,radosavovic2024humanoid}
to manipulation \citep{cheng2024expressive,he2024learning,dao2023sim,fu2024humanplus}. Dynamic humanoid locomotion has been demonstrated such as walking on rough terrain \citep{siekmann2021blind,gu-RSS-24}, resisting large disturbances \citep{van2024revisiting}, running \citep{crowley2023optimizing}, and parkour \citep{zhuang2024humanoid}. These works often utilize complex neural networks and training pipelines for high expressiveness or require a history of state-action pairs for online adaptation, reducing the sim-to-real gap in deployment. In comparison, performing dynamic motions with a simple algorithm and architecture remains challenging. Furthermore, prior works often include wide distributions of domain randomization due to the higher robustness requirement to counteract the imprecise models with complex transmissions. However, excessive randomization may hinder successful policy learning or lead to exceedingly conservative policies~\citep{chebotar2019closing}. Despite the progress in full-scaled humanoid robots, learning control policies for smaller-scale humanoids pose different challenges due to the shorter-legged design as discussed in Sec.~\ref{sec:intro}. 
Prior works, such as teaching miniature humanoid robots to play soccer, address these challenges with large flat foot designs and servo motors \citep{haarnoja2024learning}, resulting in limited dynamic motion capabilities. In contrast, our design uses smaller flat feet and more powerful actuators, enabling more dynamic motions but presenting greater control challenges.

\begin{table}[t]
\centering
\caption{Comparison of existing electric humanoid locomotion research platforms. }
\label{tab:compare_to_other_robots}
\resizebox{\linewidth}{!}{%
\begin{threeparttable}
\begin{tabular}{c|ccccccccccc}
\toprule
\multirow{2}{*}{Robot} &\multirow{2}{*}{Size\tnote{a}} & Avg. Leg \tnote{b} & Leg & Weight & Price    & Actuator\tnote{c} & Max HFE  & Max KFE  & Transmission & T/F    \\
                                     &   & Len.(m)    & Dof & (kg) & (USD) & Type  & Tor.(Nm) & Tor.(Nm) & Complexity & Sensor \\
\midrule    
TORO \citep{englsberger2014overview} & F & $\sim$0.4  & 6   & 76.4 & -     & H     & 100      & 130      & ++   & Joint     \\
LOLA \citep{seiwald2021lola}         & F & $\sim$0.44 & 6   & 68.2 & -     & H     & 370      & 390      & +++  & Feet      \\
WALK-MAN \citep{tsagarakis2017walk}  & F & $\sim$0.38 & 6   & 132  & -     & H     & 270-400  & 270-400  & ++   & Feet      \\
Unitree H1 \citep{unitreeh1}         & F & $\sim$0.4  & 5   & 47   & 90K   & P     & 270      & 360      & +    & \ding{55} \\
Digit \citep{digit}                  & F & $\sim$0.5  & 6   & 50   & 250K  & C, H  & 200      & 230      & +++  & \ding{55} \\
ARTEMIS \citep{zhu2023design}        & F & $\sim$0.38 & 5   & 37   & -     & P     & 250      & 250      & +    & Feet      \\
Cassie \citep{cassie}                & F & $\sim$0.5  & 5   & 35   & 250K  & C, H  & 195      & 195      & +++  & \ding{55} \\
MIT \citep{saloutos2023design}       & M & $\sim$0.28 & 5   & 24   & -     & P     & 72       & 144      & +    & \ding{55} \\
Unitree G1 \citep{unitreeg1}         & M & $\sim$0.3  & 6   & 35   & 16K   & P     & 88       & 139      & +    & \ding{55} \\
HECTOR \citep{li2023dynamic}         & M & $\sim$0.22 & 5   & 16   & -     & P     & 33.5     & 51.9     & +    & \ding{55} \\
iCub \citep{parmiggiani2012design}   & M & $\sim$0.2  & 6   & 24   & 300K  & H     & 40       & 40       & ++++ & Feet      \\
BRUCE \citep{liu2022design}          & S & $\sim$0.17 & 5   & 3.3  & 6.5K  & P     & 10.5     & 10.5     & +    & \ding{55} \\
NAO \citep{gouaillier2009mechatronic}& S & $\sim$0.15 & 6   & 4.5  & 14K   & S     & 1.61     & 1.61     & +    & Feet      \\
DARwIn-OP \citep{ha2011development}  & S & $\sim$0.09 & 6   & 2.8  & -     & S     & 2.35     & 2.35     & +    & Feet      \\
Surena-Min \citep{nikkhah2017design} & S & $\sim$0.085& 6   &  3.3 & -     & S     & 3.1      & 7.3      & +    & \ding{55} \\
\textbf{Ours}                        & M & \textbf{$\sim$0.2} & \textbf{6} & \textbf{16\tnote{d}} & \textbf{10K\tnote{e}} & \textbf{P} & \textbf{62.6} & \textbf{81.1} & \textbf{+} & \ding{55} \\
\bottomrule
\end{tabular}
\begin{tablenotes}
    \item[a] F, M, and S represent Full, Middle, and Small, respectively.
    \item[b] Average length of thigh and calf.  
    \item[c] \parbox[t]{0.72\linewidth}{H, P, C, and S represent Harmonic Drive, Planetary, Cycloidal Drive, and Servo Motor with a high reduction ratio, respectively.}
    \item[d] Without arms. The estimated weight of two 4 DoF arms is 6kg, the total weight will be 22kg.
    \item[e] Without arms. The estimated cost of two 4 DoF arms is 5K USD, the total non-profit cost will be 15K USD.
\end{tablenotes}
\end{threeparttable}
}
\end{table}

\section{Design for Learning-based Control}
\label{sec:hardware}

\begin{figure}[t]
    \centering
    \begin{tikzpicture}[inner sep=0pt]
        \def\height{0.33\linewidth}
        \node (component) {\includegraphics[page=1,height=\height,trim={150 0 125 0},clip]{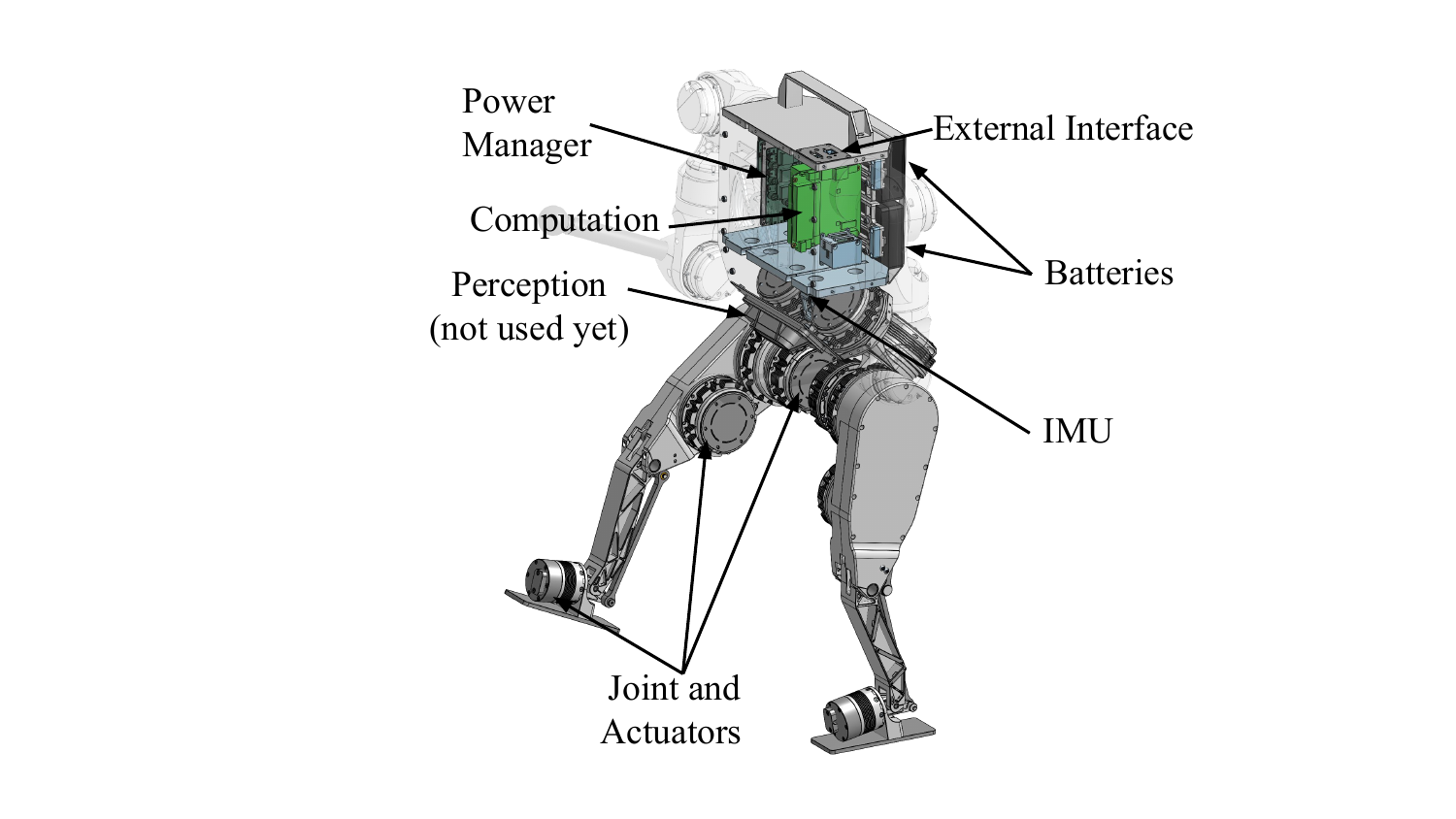}};
        \node[right=0pt of component] (joints) {\includegraphics[page=2,height=\height,trim={200 0 200 0},clip]{figures/figures_from_slides.pdf}};
        \node[right=0pt of joints] (actuators) {\includegraphics[page=3,height=\height,trim={150 0 160 20},clip]{figures/figures_from_slides.pdf}};
        \node[black, anchor=south west, inner sep=5pt] at (component.south west) {(a)};
        \node[black, anchor=south west, inner sep=5pt] at (joints.south west) {(b)};
        \node[black, anchor=south west, inner sep=5pt] at (actuators.south west) {(c)};
    \end{tikzpicture}
    \caption{Overview of design: (a) main components, (b) joints and key dimensions, (c) key actuators and joints of the left leg.}
    \label{fig:overview}
\end{figure}

\begin{figure}[t]
    \centering
    \begin{tikzpicture}[inner sep=0pt]
        \def\height{0.27\linewidth}
        \node(exposed) {\includegraphics[page=5,height=\height,trim={50 50 50 50},clip]{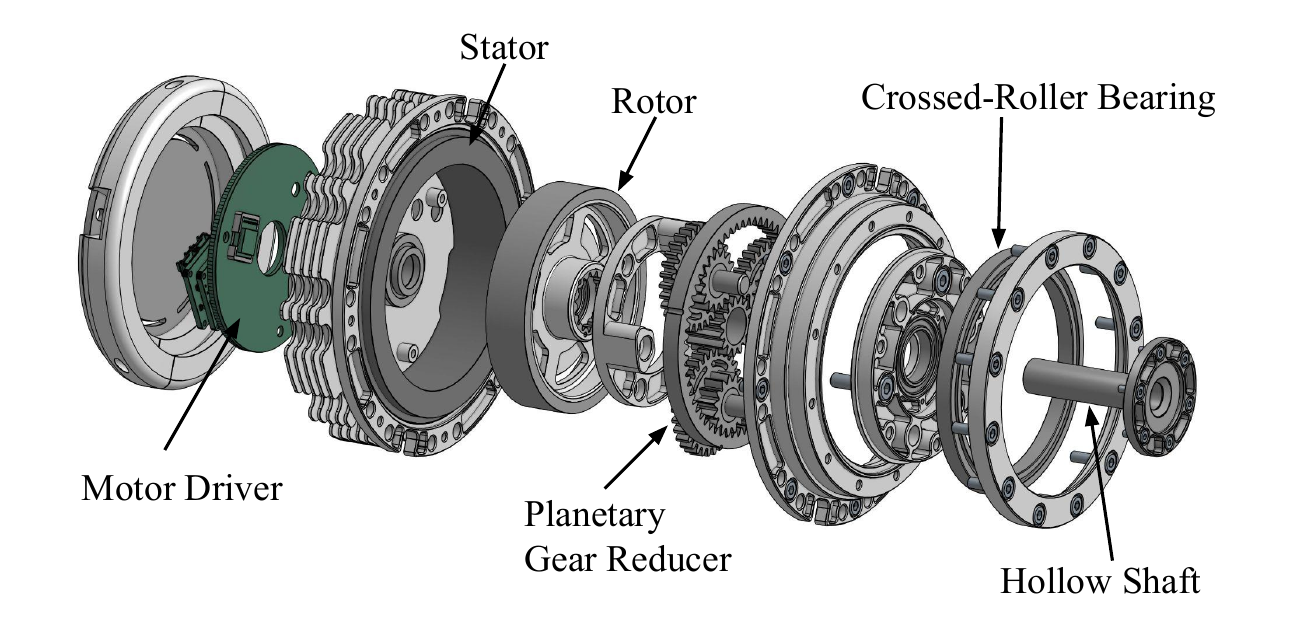}};
        \node[right=0pt of exposed] (cross) {\includegraphics[page=6,height=\height,trim={40 0 50 0},clip]{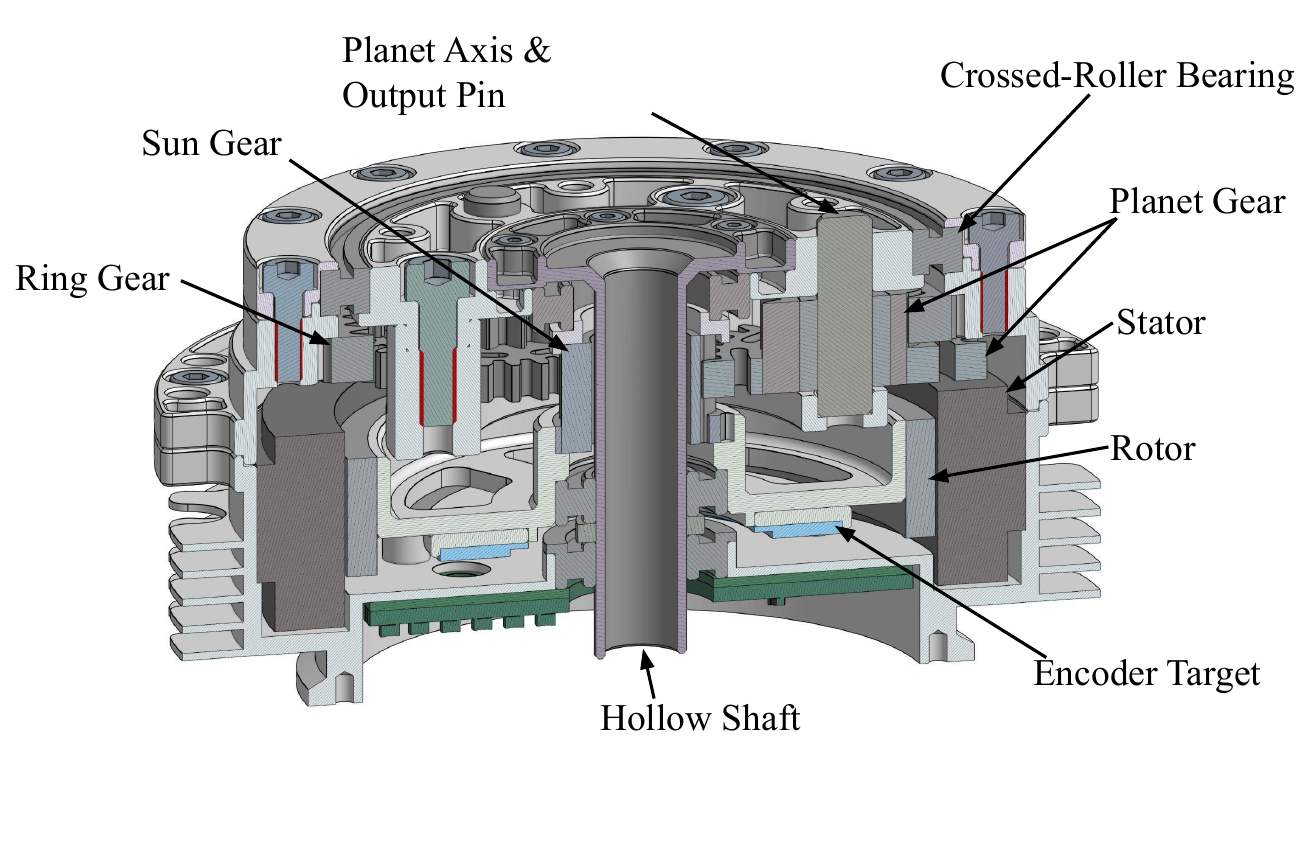}};
        \node[black, anchor=south west, inner sep=5pt] at (exposed.south west) {(a)};
        \node[black, anchor=south west, inner sep=5pt] at (cross.south west) {(b)};
    \end{tikzpicture}
    \caption{(a) Exposed view and (b) cross view of one of our custom actuators.}
    \label{fig:actuator_details}
\end{figure}

In this section, we will introduce our humanoid robot design. First, we provide an overview of the system design, and then, we will explain the motivations and our solutions behind the design choices tailored for learning-based control algorithms. 

\subsection{System Overview}
The Berkeley Humanoid is a 16 kg, fully electric drive mid-scale robot for humanoid research. The main component is shown in \figref{fig:overview}(a). The robot has a torso and two 6 DoF legs, with a thigh length of 220 mm, a calf length of 180 mm, and a total height of 0.85 m in a nominal standing configuration, resembling a 5-year-old child in body shape.

Inside the torso, a computer, a power management board, and a cheap cell phone level IMU sensor are installed. Besides these, two easily changeable batteries are mounted in a protected compartment in the torso. Each leg is equipped with 6 actuators for the 6 joints, most of which are directly attached to the link and act as a joint. 
Two 4-DoF arms were designed but left out to simplify and focus on locomotion abilities in this work. 
To adapt to different torque requirements on each joint, we built 4 types of actuators named according to the motor size and 2 types of motor drivers for each leg, as shown in \tabref{tab:actuator_spec} and \figref{fig:actuators_in_robot}. These high-performance actuators allow our robot to perform highly dynamic maneuvers.

The communication system is another critical component of our robot design. To enable accurate communication with minimum latency, we opt for high-bandwidth EtherCAT protocol.
We develop custom EtherCAT clients for both custom motor drivers and the IMU. The onboard PC runs the EtherCAT master and communicates with the peripherals at frequencies ranging from 1 kHz to 4 kHz. USB and ethernet connection is also supported for the externally interfacing sensors, such as depth camera, lidar, or other sensors. For the user development and debugging interface, a router inside the torso provides both wire and wireless connections to the onboard PC.

Almost every part of the robot is custom-designed and built, including the actuators, mechanical components, motor driver, IMU, communication, and power management board. 
This comprehensive understanding of the whole system enables us to explore new control strategies with a narrow sim-to-real gap, as well as accounting for specific requirements on the hardware from learning-based algorithms, namely, being \textbf{Simulation-Friendly}, \textbf{Reliable and Low-cost}, \textbf{Experiment-Friendly}, and \textbf{Anthropomorphic}. We will provide more detail on each of these next.

\begin{table}[t]
\centering
\caption{Custom Actuator Specifications.}
\label{tab:actuator_spec}
\begin{tabular}{c|cccc}
\toprule
Actuator                       & 5013          & 8513        & 8518        & 10413 \\
\midrule
Mass (g)                       & 251           & 756         & 856         & 1011         \\
Gear Ratio                     & 9:1           & 9:1         & 9:1         & 9:1          \\
Hollow Shaft                   & \ding{55}     & \ding{51}   & \ding{51}   & \ding{51}    \\
Diameter × Thickness (mm)      & 54.6 × 53     & 104 × 50    & 104 × 55    & 123 × 50     \\
Peak Torque (Nm)               & 9.7           & 45.3        & 62.6        & 81.1         \\
Sustained Torque (Nm)          & 4.59          & 18.9        & 26.1        & 34.2         \\
Max. Speed at 48V (rad/s)      & 83.7          & 40.7        & 29          & 27.9         \\
Max. Power (W)                 & 220           & 570         & 730         & 890          \\
Rotor Inertia ($\text{kgm}^2$) & 6.1e-6        & 6.9e-5      & 9.4e-5      & 1.5e-4       \\
\bottomrule
\end{tabular}
\end{table}

\subsection{Simulation-Friendly}
\paragraph{Motivation.}
Since the dominant trend of modern learning-based locomotion policies leverages model-free reinforcement learning with massively parallelizable simulators as the learning platform, a key consideration of our robot is its simulation cost. For example, while designing transmission linkages with unilateral springs may reduce the load for joint motors, and absorb large impacts, the resulting mechanism involves solving extra dynamical equations that are notoriously hard to simulate and result in high computation costs for parallelism. Furthermore, as most simulators typically model robots with multi-rigid-body dynamics, some can only apply torque directly in joint space without considering actuator transmissions, while others require much more computation to solve the closed kinematic chains involved in the transmissions. 
However, actuator and transmission factors that can significantly alter the actuation dynamics during highly dynamic tasks, such as torque, velocity, position limits, sensor noise, friction, and inertia of the linkage and rotor, are very challenging to accurately and efficiently map and randomize in joint space.
Additionally, more computation and smaller timesteps are required to simulate communication delays \cite{tan2018sim,li2021reinforcement,xie2020learning}, motor/actuator dynamics \cite{hwangbo2019learning}, and inaccurate execution rates \cite{siekmann2021blind}, which further slows down the simulation.

\paragraph{Our Approach.}
To avoid these difficulties, we opt to remove all flexible or energy-absorbing components, such as springs or dampers, as well as any closed kinematic chains from the robot's kinematic chain and use the simplest actuator-joint transmissions.
As illustrated in \figref{fig:actuator_details}, all actuators are equipped with a cross roller bearing, so that the actuators can be directly mounted and used as joints. As a result, rotor inertia can be easily simulated by adding armature to the diagonal of the joint mass matrix, and other actuator factors can be modeled the same as the joint.
One exception is the FFE joint shown in \figref{fig:overview}, where a linkage transmission is employed to provide large torques, resulting in a coupled but linear joint-actuator mapping for KFE and FFE. This design allows us to treat the actuator as a joint in simulation. 
In addition, the selection of a planetary gearbox with a QDD gear ratio in our actuators introduces only minor friction uncertainties which are easy to model in joint space. By combining these designs during training, we can focus solely on joint simulation without considering actuator dynamics. 
To avoid simulating system latency, we use EtherCAT for communications. This ensures a negligible maximum latency ranging from 0.5 ms to 2 ms\footnote{The exact latency depends on the selected frequency: 2 ms at 1 kHz and 0.5 ms at 4 kHz.}. The motor torque control bandwidth is set to 1 kHz, allowing the actuator to be simulated as a torque source without delay.
These designs enable our robot to achieve an accurate simulation at an efficiency of more than 90,000 simulation steps per second on an NVIDIA A4500 GPU.

\subsection{Reliable and Low-Cost}
\begin{table}[t]
\centering
\caption{Cost of Each Component in Small Quantity Production.}
\label{tab:cost}
\begin{tabular}{c|cccc|c|cc|cc|c}
\toprule
\multirow{2}{*}{Module} & \multicolumn{4}{c|}{Actuator} & Sensor &  \multicolumn{2}{c|}{Misc} & \multicolumn{2}{c|}{Off-the-shelf} & \multirow{2}{*}{Total}\\
                  & 5013 & 8513 & 8518 & 10413 &  IMU & Torso & Leg & PC  & Battery \\
\midrule
Cost (USD)        & 422  & 570  & 639  & 676   & 50   & 410  & 974  & 347 & 153 & 9955 \\
Quantity          & 2    & 6    & 2    & 2     & 1    & 1     & 2   & 1   & 2   &- \\
\bottomrule
\end{tabular}
\end{table}

\paragraph{Motivation.}
In the past, humanoid locomotion research required high-end robots, accurate sensors, careful protection, and lengthy repairs, limiting the field's development. 

In order to accelerate the field further and to make a change, our robot must be reliable and accessible, meaning that it should be durable for repeated experiments and of low cost. A more accessible and reliable robot also paves the way for scaling up humanoid robot learning in real-world settings.

\paragraph{Our Approach.}
In order to improve durability, we build the robot with high-performance materials as opposed to \cite{kscale, grimminger2020open, urs2022design, song2022drpd, fuge2023design}. We use 7075 and 6061 aluminum for building most of the main components, and SKD11 steel for the gearbox and linkage, allowing the robot to survive heavy impacts with lightweight structures. The endurance of electrical cables for power and signals is a key factor for the reliability of the robot, where contact with the environment creates tearing due to friction and vibrations that post significant challenges for cable durability. To overcome this, we opt to leverage hollow shaft designs for most of the actuators as shown in \tabref{tab:actuator_spec}, where power and communication cables cross between the two moving bodies through the hollow shaft axis of the joint, minimizing the tearing caused by joint movements. Furthermore, the usage of custom QDD actuators allows us to estimate the joint torque without adding strain gauges. With reliable joint torque sensing, a generalized momentum observer \cite{haddadin2008collision} can be used to estimate the contact wrench of each foot without requiring contact sensors or force/torque sensors, which further improves the reliability of the robot.

The fully customized hardware allowed us to minimize the robot's cost, as shown in \tabref{tab:cost}. 
With learning algorithms, we typically gain enhanced robustness against hardware inaccuracies, allowing for cheaper sensors and further cost reductions. Thus, unlike most previous works \cite{katz2018low,spot,cassie,digit} where the IMU costs around USD 1,000, we can utilize a cell phone level IMU ICM42688 that costs less than one dollar\footnote{For sensor IC itself, net cost of IMU Module shown in \tabref{tab:cost}.}. These designs help cut the cost down to USD 10,000 for the whole robot without arms.
Note that most costs shown in \tabref{tab:cost} will decrease with scaled-up production. The only non-custom components are the computers (Intel i7-1255U) and batteries (DJI TB50), sourced commercially for performance and safety.

\subsection{Experiment-Friendly}
\paragraph{Motivation.}
In the past, the size and weight of humanoid robots are especially troublesome for experiments. Traditional full-scaled humanoids are often heavier than a person of the same size, which means handling the robot requires at least two or three people with the help of gantries. More importantly, experimenting with such robots with high torque actuators ($\approx$300 Nm) is dangerous and may result in severe injuries to people nearby. 

\paragraph{Our Approach.}
By properly choosing the robot size and the custom lightweight materials, we reduced the weight to only 16 kg, which allows us to do experiments with only one robot operator for indoor environments, and with an optional cameraman in outdoor environments, including commanding the robot, collecting data, taking video, and sometimes resetting the robot from failure. All of the experiments reported in this work are done with this setup.

\subsection{Anthropomorphic}
\label{sec:anthropomorphic}
\paragraph{Motivation.}
The advantage of using an anthropomorphic design is significant: it allows for higher static stability and human-like motions by having similar dominant DoFs as human bodies. This results in wider applicability, richer task selection, and easier learning from widely available human demonstrations. 

\paragraph{Our Approach.}
The dominant motion of a human leg \cite{kudo2023optimal}, while we can model a foot contact with the ground as a 6 DoF contact wrench \cite{caron2015stability}. 
Our robot uses an anthropomorphic design with 6 DoFs per leg, which replicates the common modeling of DoFs human legs have. Compared to~\cite{unitreeh1, cassie,li2023dynamic, zhu2023design, saloutos2023design}, providing actuation on the roll direction of the ankle joint improves the robot's stability in challenging static poses, such as when manipulating distant objects, and enables it to potentially balance on one foot. Furthermore, each joint limit is designed to closely align with the corresponding physical limits of human bodies. This allows us to provide further protection on the hardware while ensuring enough ranges for imitating human motions. 

\section{A Minimally Composed Learning-based Controller}
\label{sec:control}

With a humanoid platform designed for learning-based control, we are able to achieve robust and agile locomotion with a minimally composed RL controller. In this section, we first introduce the design of the RL controller. Then, we elaborate on how our humanoid platform enables the narrowing of the sim-to-real gap for the RL controller. 

\subsection{Reinforcement Learning Formulation}
We formulate our tasks as Markov Decision Processes (MDPs) and leverage RL to solve them due to their promising performance in humanoid control.
We create a minimally composed learning-based controller by doing the following.  We formulate the MDP with minimal observation and action spaces. Specifically, we only use immediate state feedback as actor input, without formulating a short or long history~\cite{siekmann2021blind,li2024reinforcement} or teacher-student training~\cite{kumar2021rma,lee2020learning} to estimate environment parameters. 
Similarly, we opt out of pre-defined phase signals~\cite{siekmann2021sim} or reference motion~\cite{li2024reinforcement} to reduce human biases. 
The immediate state feedback includes raw proprioceptive readings (base angular velocity $\mathbf{\omega}$, projected gravity vector $\mathbf{g}$, joint positions $\mathbf{q}$, velocities $\mathbf{\dot{q}}$), base linear velocity $\mathbf{v}$ from a state estimator~\cite{flayols2017experimental}, velocity commands $\mathbf{v}^c_{x,y}$ and $\mathbf{\omega}^c_z$, and the previous action.
Likewise, the action space consists solely of the desired joint positions $\mathbf{q}^d$, which are converted into torques $\mathbf{\tau}$ directly by a PD controller on the motor driver.

We also design the architecture of the actor-critic with the most basic multilayer perceptron (MLP) networks only. Specifically, each network has hidden sizes of $[512, 256, 128]$ neurons and ELU activation. The policy is optimized via PPO~\cite{schulman2017proximal} and trained in Isaac Lab \citep{mittal2023orbit}. The RL policy executes at 50 Hz, the state estimator at 1 kHz, and the PD controller at 25 kHz. 

This minimally-composed RL controller facilitates the validation of the adequacy of our hardware design for learning-based control. Without the ability to do online system identification (through the I/O history) or reference motion guidance, our policy relies on the synergy of the hardware and learning algorithm to achieve a narrow sim-to-real gap, ensuring that the robust and agile locomotion performance in training can be fully demonstrated on the real-world robot. Additionally, it serves as a competent baseline for other algorithms developed on our platform.

\subsection{Closing the Sim-to-Real Gap}
\label{sec:sim2real}

\paragraph{Hardware Side.}
We focus on closing the sim-to-real gap through hardware design choices. The main factors of the sim-to-real gap, aside from sensor noise, are modeling errors and command execution rate, accuracy, and delay~\cite{xie2020learning,tan2018sim,hwangbo2019learning}. 
To reduce modeling errors, we install actuators directly as joints or design a linear joint-actuator mapping, avoiding the simulation of structures that are likely to result in inaccurate modeling.
To improve command execution, we employ high bandwidth torque control that leads to a precise execution rate, and transparent QDD actuator dynamics so that the commanded torque is accurately tracked and has negligible communication latency. All of these lead to less discrepancy between the hardware and the simulated dynamics.

\paragraph{Design-enabled Accurate Domain Randomization.}

While most of the learning controllers rely on domain randomization, extensive domain randomization slows down training and results in conservative policies~\cite{chebotar2019closing}. To avoid this while still preserving a robust policy, in this work, we leverage a different approach aimed at providing accurate domain randomizations given the hardware design. For a humanoid robot performing locomotion tasks, we identify two sources of uncertainties: uncertainty in the robot physics property, e.g., the mass of each link, and that in performing tasks, e.g., contact with the environment. 

For hardware uncertainty, our detailed design allows us to obtain a small and accurate range of parameter variations. Specifically, we use CAD to retrieve accurate mechanical parameters like rotor inertia and conduct simple experiments to characterize the friction of each actuator separately. This demonstrates the benefits of an in-house-built robot, as obtaining such detailed hardware parameters for commercial robots would be difficult.

For uncertainty in contact with the environment, we apply a wide range of domain randomization to cover as many real-world environment conditions as possible. This includes ground friction, restitution, and external perturbation forces from obstacles and unstable ground conditions. 

Unlike previous work~\citep{tan2018sim, peng2020learning, li2024reinforcement}, we opt not to randomize properties that cannot be identified in these two categories, such as a general ``motor strength" ratio or PD gains, which were often used as a ``lazy approach" to approximate actuation uncertainties. However, because it is hard to accurately analyze the range of uncertainties with PD approximation, prior works rely on heuristics, which can lead to unnecessarily large ranges of domain randomization, which we aim to avoid. %

As we will show later, with design-enabled accurate domain randomizations, we can achieve robust and agile locomotion skills when zero-shot transferring to robot hardware, even with a minimally-composed RL controller. 

\section{Experimental Validation}
\label{sec:experiment}
In our experiments, we aim to validate how our humanoid design facilitates learning locomotion control from three aspects:
(1) The effectiveness of our minimally-composed RL controller in learning humanoid locomotion tasks. (2) The sim-to-real gap for the minimal RL algorithm with our adequate hardware design. 
(3) The hardware reliability of the robot.

\begin{figure}[t]
    \centering
    \begin{tikzpicture}[line width=1.5pt,inner sep=0pt]
    \newlength{\width}
    \setlength{\width}{0.95\linewidth}
    \def\smallgap{1.5pt}
    \node (straight) {\includegraphics[width=0.5\width]{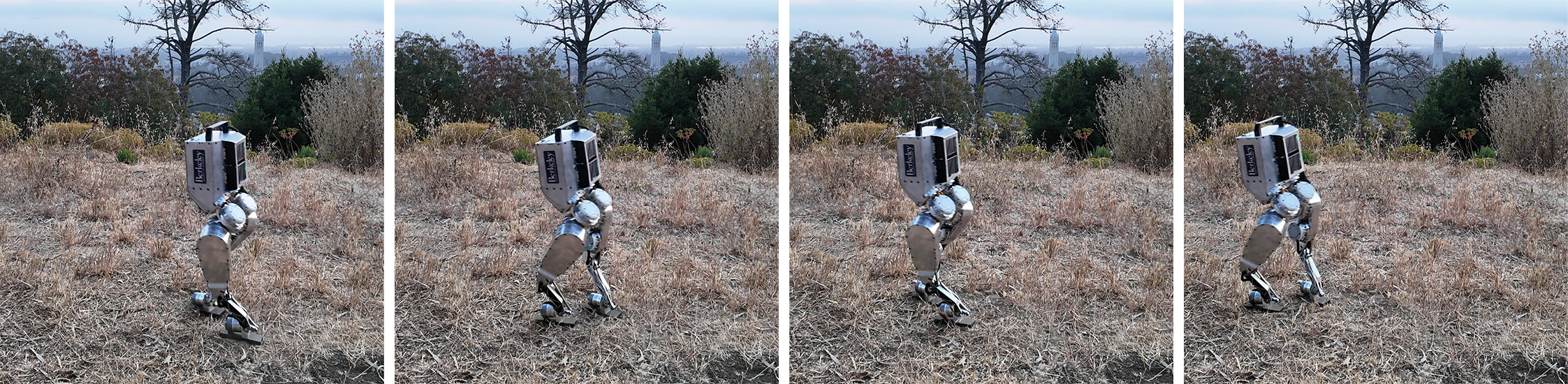}};
    \node[right=\smallgap of straight] (sideways) {\includegraphics[width=0.5\width]{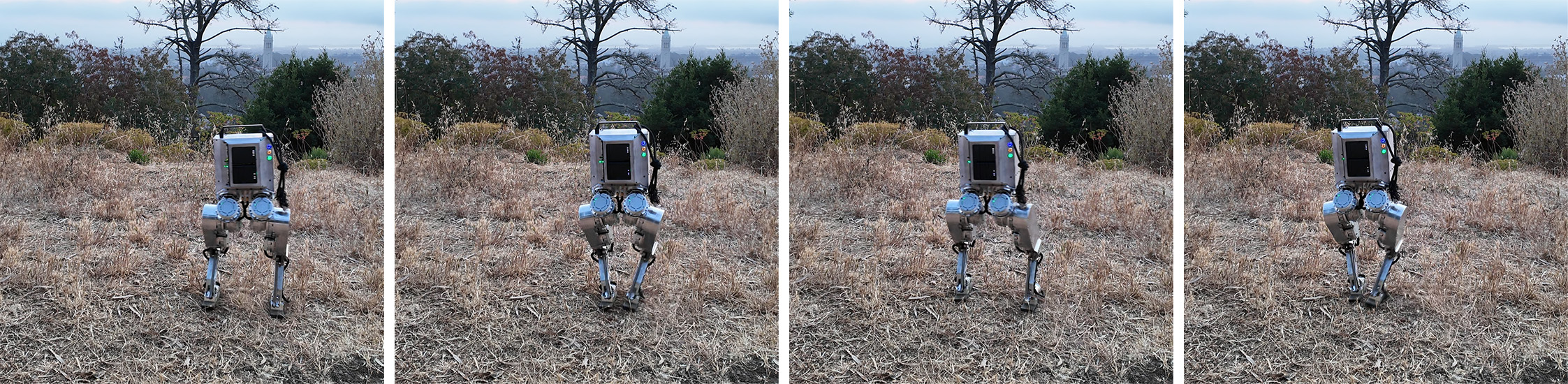}};
    \node[above=\smallgap of straight.north east] (turning) {\includegraphics[width=0.33\width]{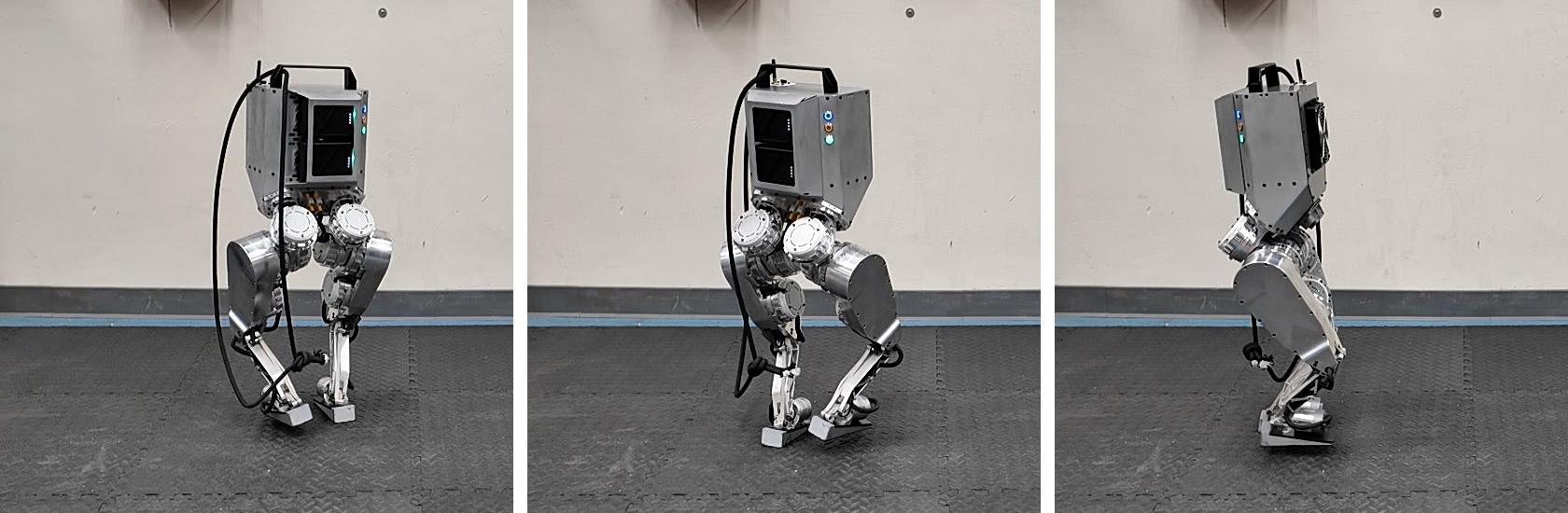}};
    \node[left=\smallgap of turning] (forward) {\includegraphics[width=0.33\width]{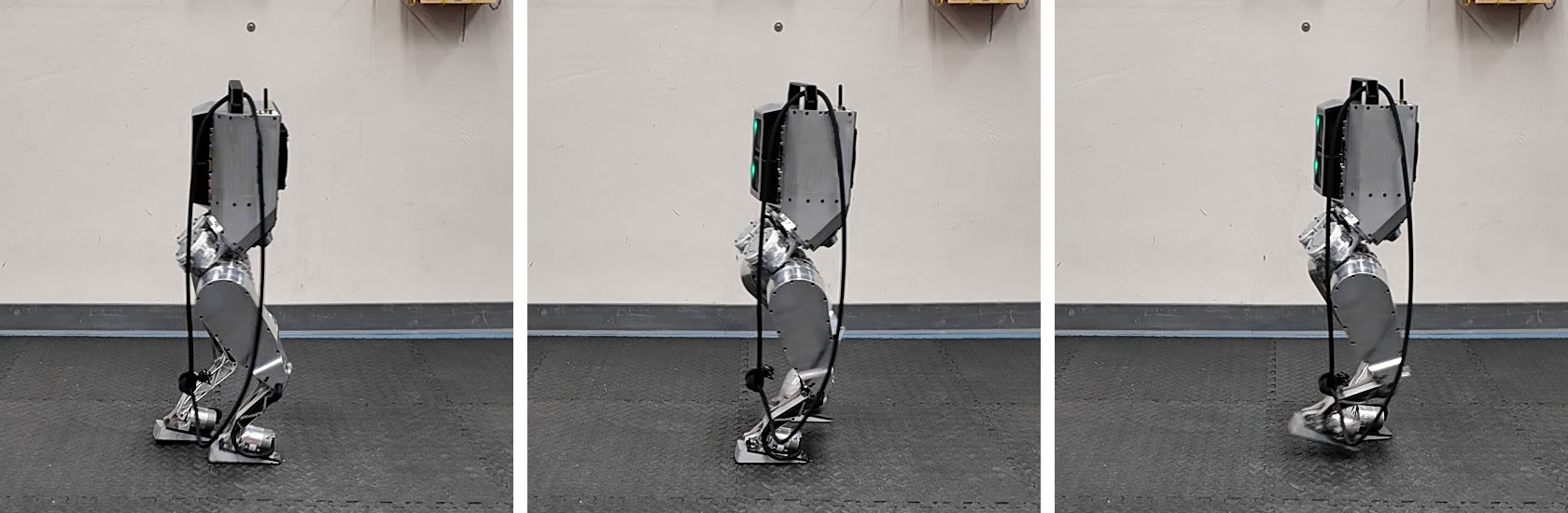}};
    \node[right=\smallgap of turning] (backward) {\includegraphics[width=0.33\width]{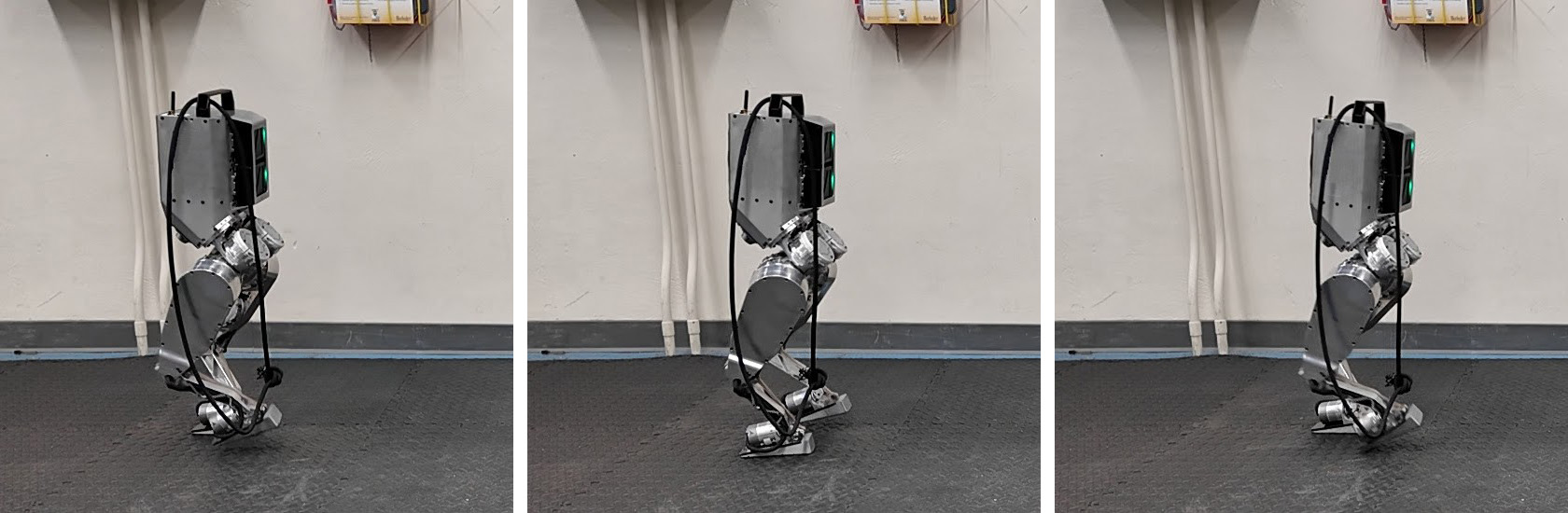}};
    \draw[my_blue] (forward.south west) rectangle (backward.north east);
    \foreach \i in {1,...,2}
    \draw[my_blue] ($(forward.south west)!\i*0.333!(backward.south east)$) -- ($(forward.north west)!\i*0.333!(backward.north east)$);
    \draw[my_yellow] (straight.south west) rectangle (sideways.north east);
    \draw[my_yellow] ($(straight.south west)!0.5!(sideways.south east)$) -- ($(straight.north west)!0.5!(sideways.north east)$);

    \node[white, rectangle, fill=my_blue, inner sep=1pt, anchor=south west, align=center] at (forward.south west) {\footnotesize (a)};
    \node[white, rectangle, fill=my_blue, inner sep=1pt, anchor=south west, align=center] at (turning.south west) {\footnotesize (b)};
    \node[white, rectangle, fill=my_blue, inner sep=1pt, anchor=south west, align=center] at (backward.south west) {\footnotesize (c)};
    \node[white, rectangle, fill=my_yellow, inner sep=1pt, anchor=south west, align=center] at (straight.south west) {\footnotesize (d)};
    \node[white, rectangle, fill=my_yellow, inner sep=1pt, anchor=south west, align=center] at (sideways.south west) {\footnotesize (e)};
    \end{tikzpicture}
    \caption{Omnidirectional Walking. (a-c) The robot walks forward, turns in place, and walks backward in the lab environment. (d, e) The robot walks forward and sideways in the wild.}
    \label{fig:omni}
\end{figure}
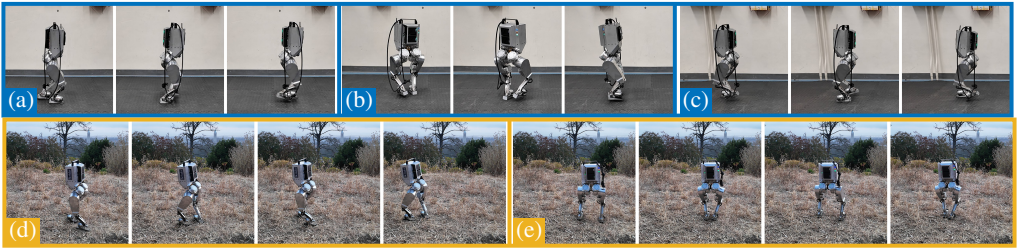

\begin{figure}[t]
    \centering
    \begin{tikzpicture}[line width=1.5pt,inner sep=0pt,outer sep=0pt]
    \let\width\relax
    \newlength{\width}
    \setlength{\width}{0.95\linewidth}
    \def\smallgap{1.5pt}

    \node (grass) {\includegraphics[width=0.25\width,trim={250 0 350 0},clip]{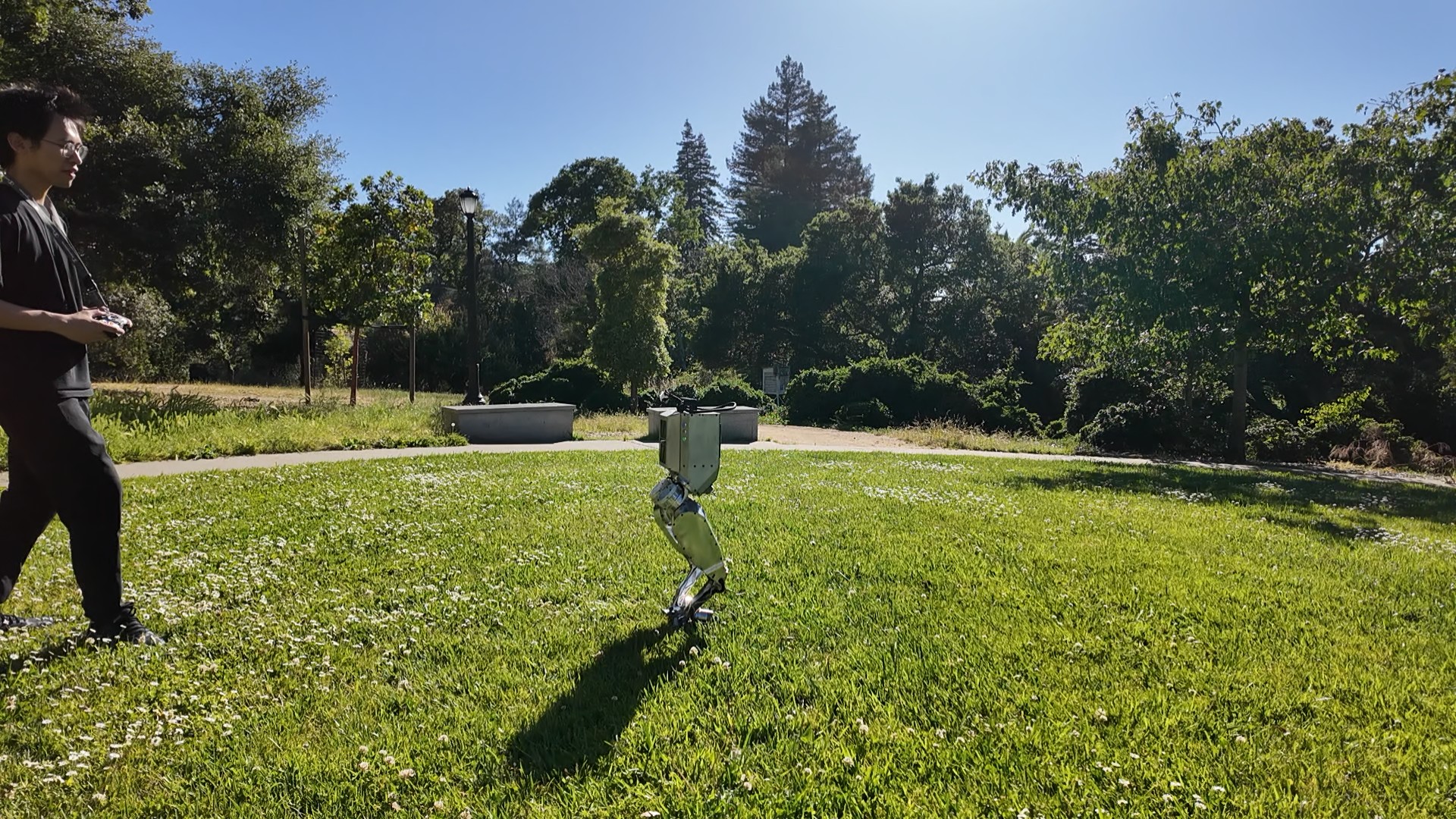}};
    \node[right=0pt of grass] (brick) {\includegraphics[width=0.25\width,trim={300 0 300 0},clip]{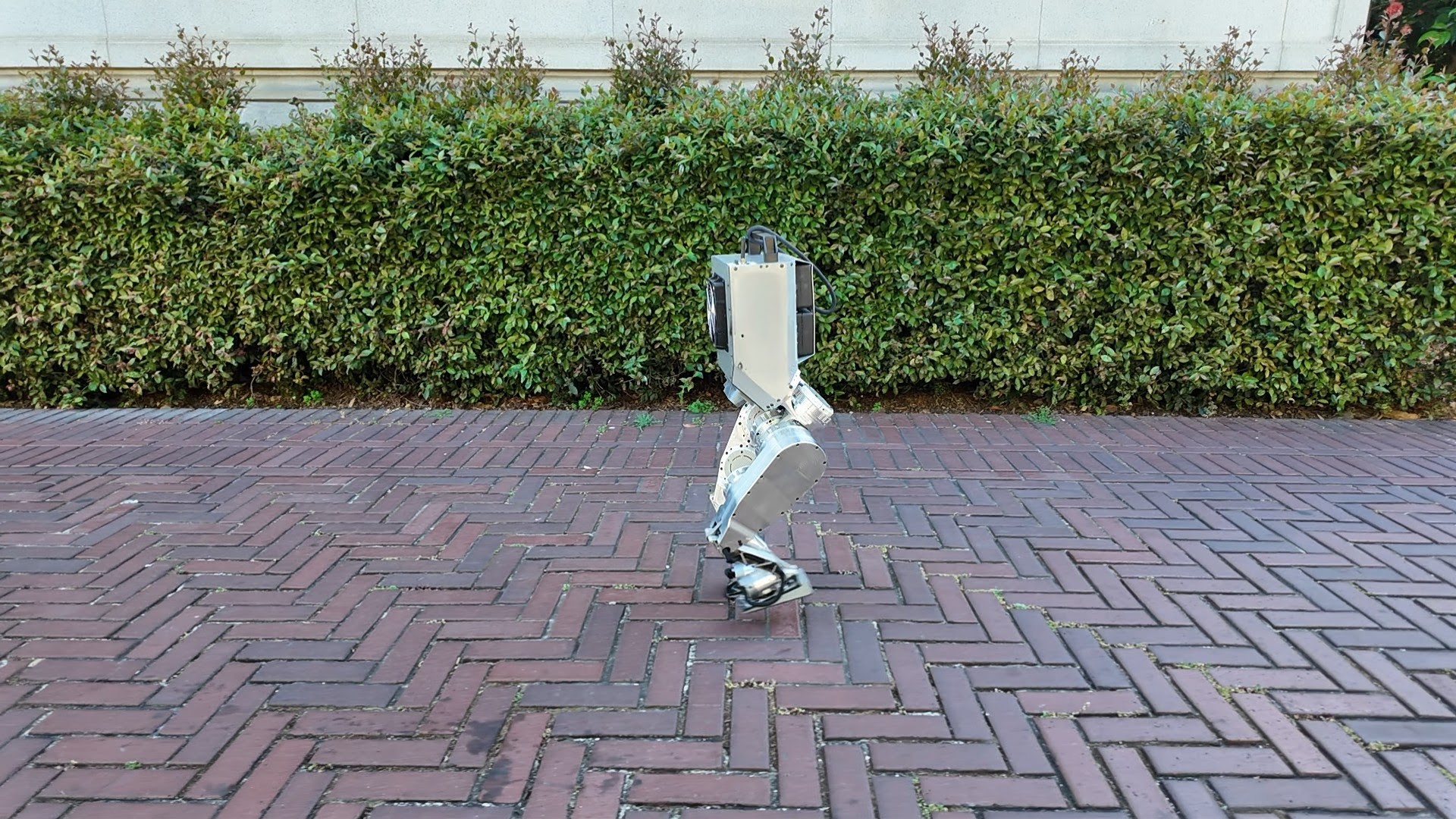}};
    \node[right=0pt of brick] (unpaved) {\includegraphics[width=0.25\width,trim={300 0 300 0},clip]{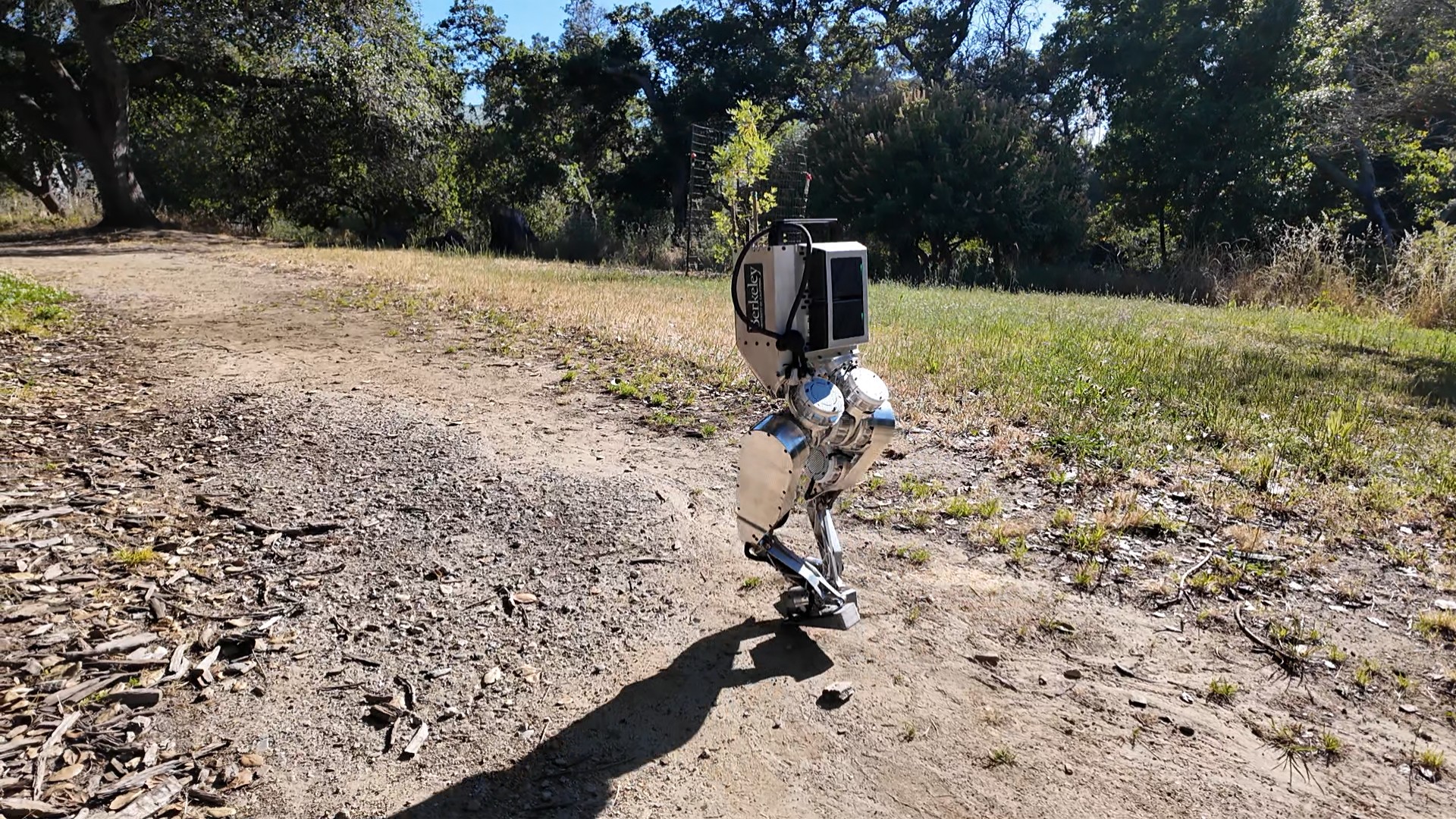}};
    \node[right=0pt of unpaved] (road) {\includegraphics[width=0.25\width,trim={300 0 300 0},clip]{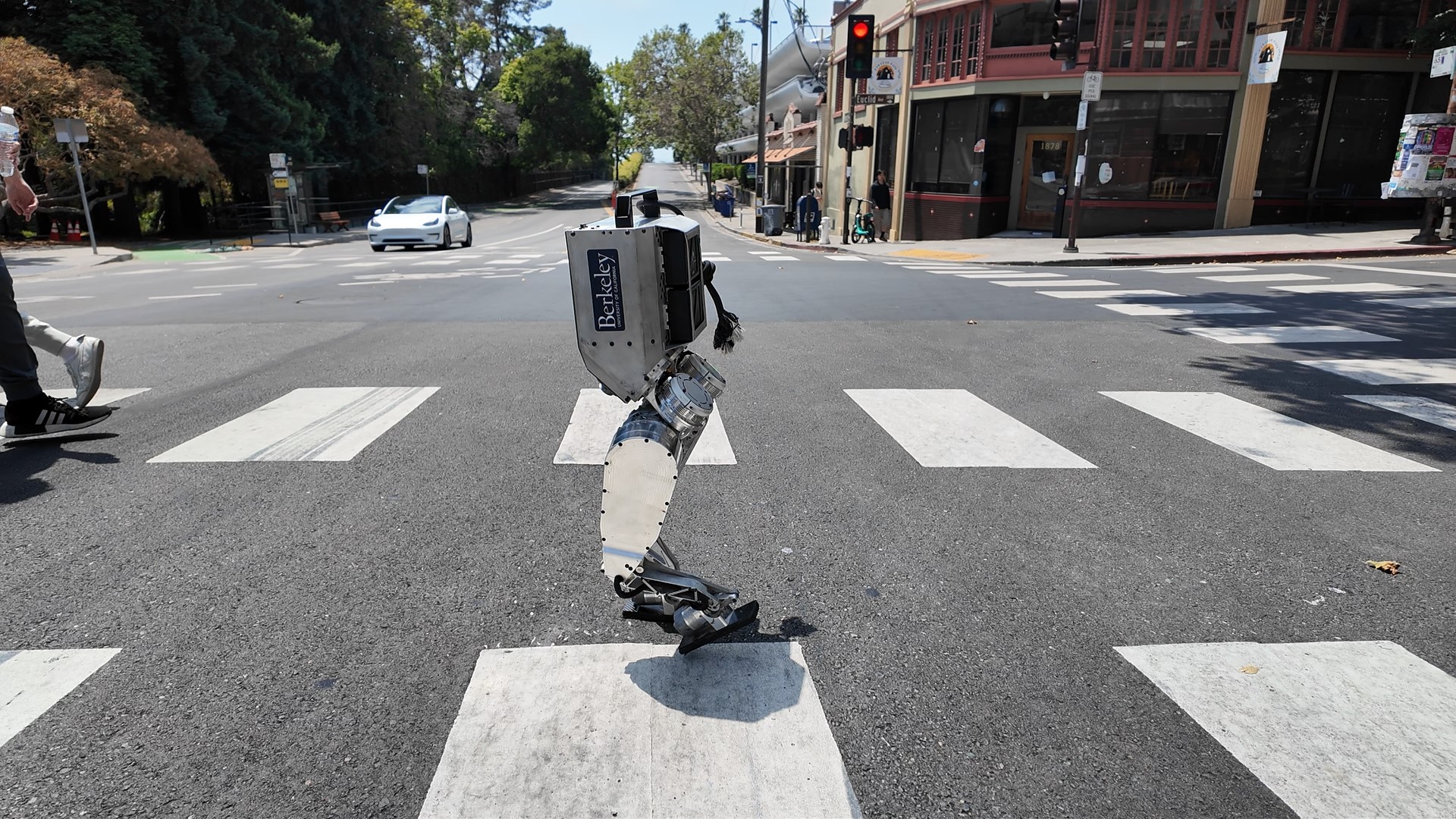}};
    \node[below=0pt of grass] (bridge) {\includegraphics[width=0.25\width,trim={300 0 300 0},clip]{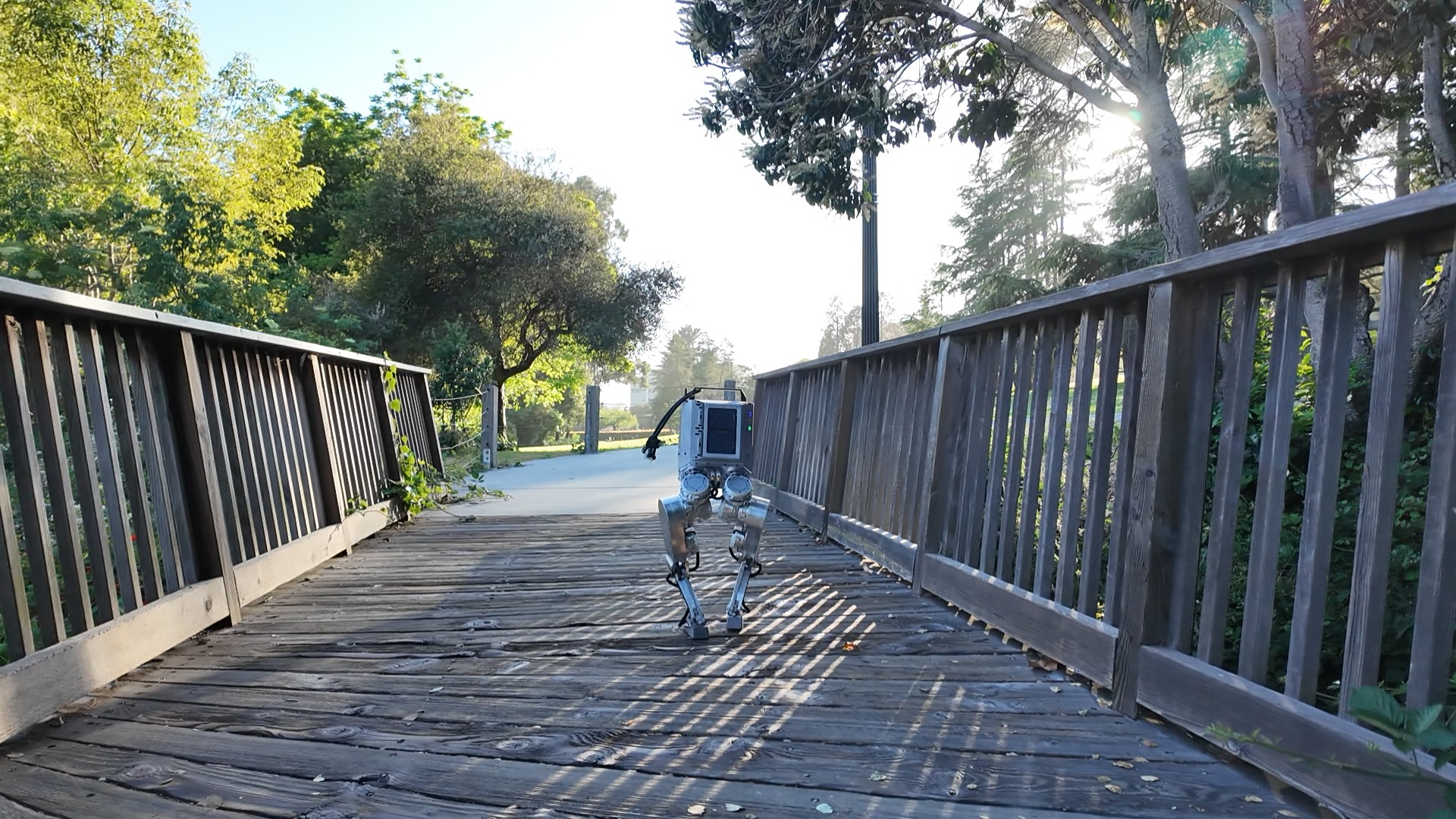}};
    \node[below=0pt of brick] (cement) {\includegraphics[width=0.25\width,trim={350 0 250 0},clip]{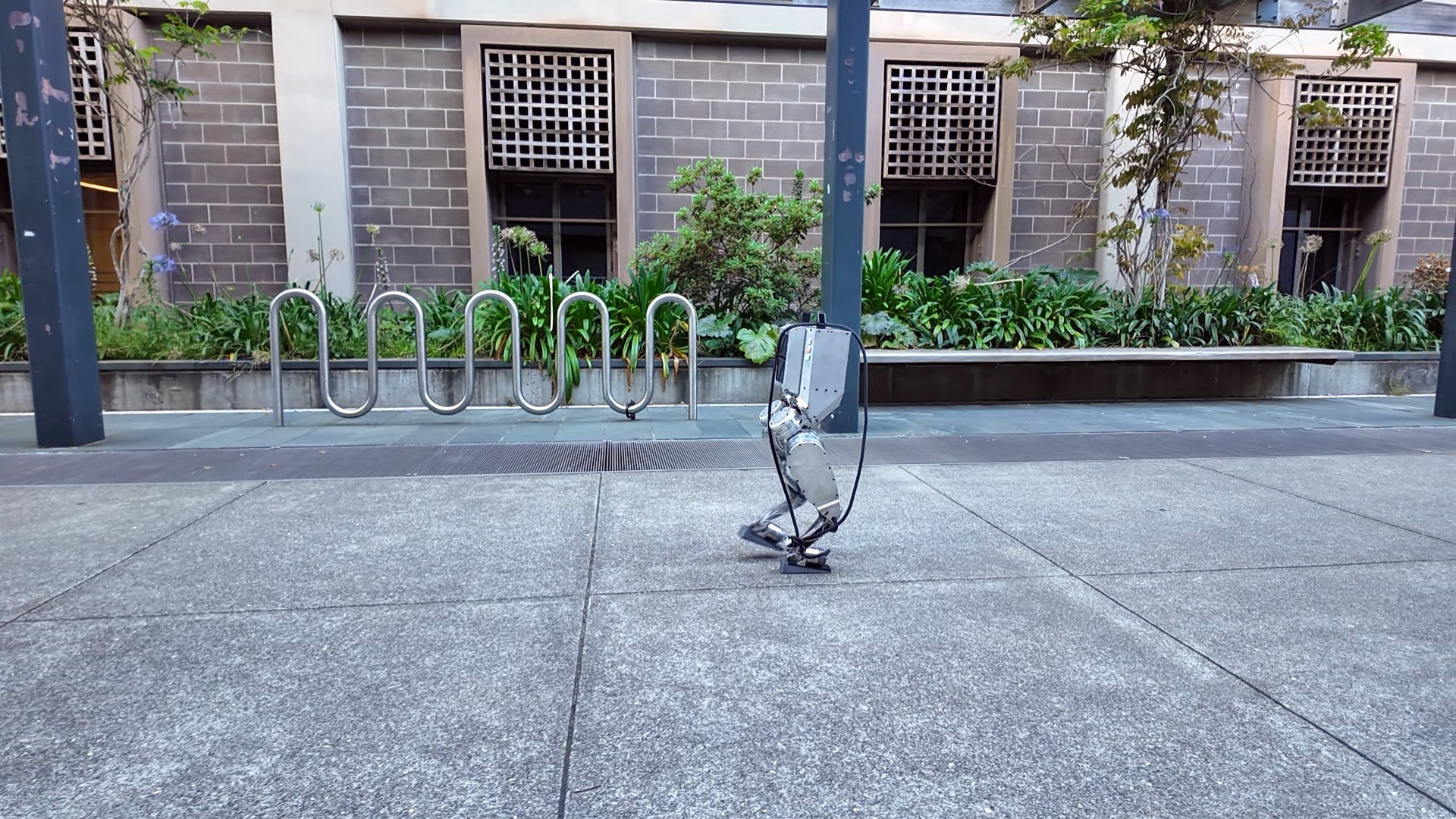}};
    \node[below=0pt of unpaved] (track) {\includegraphics[width=0.25\width,trim={350 0 250 0},clip]{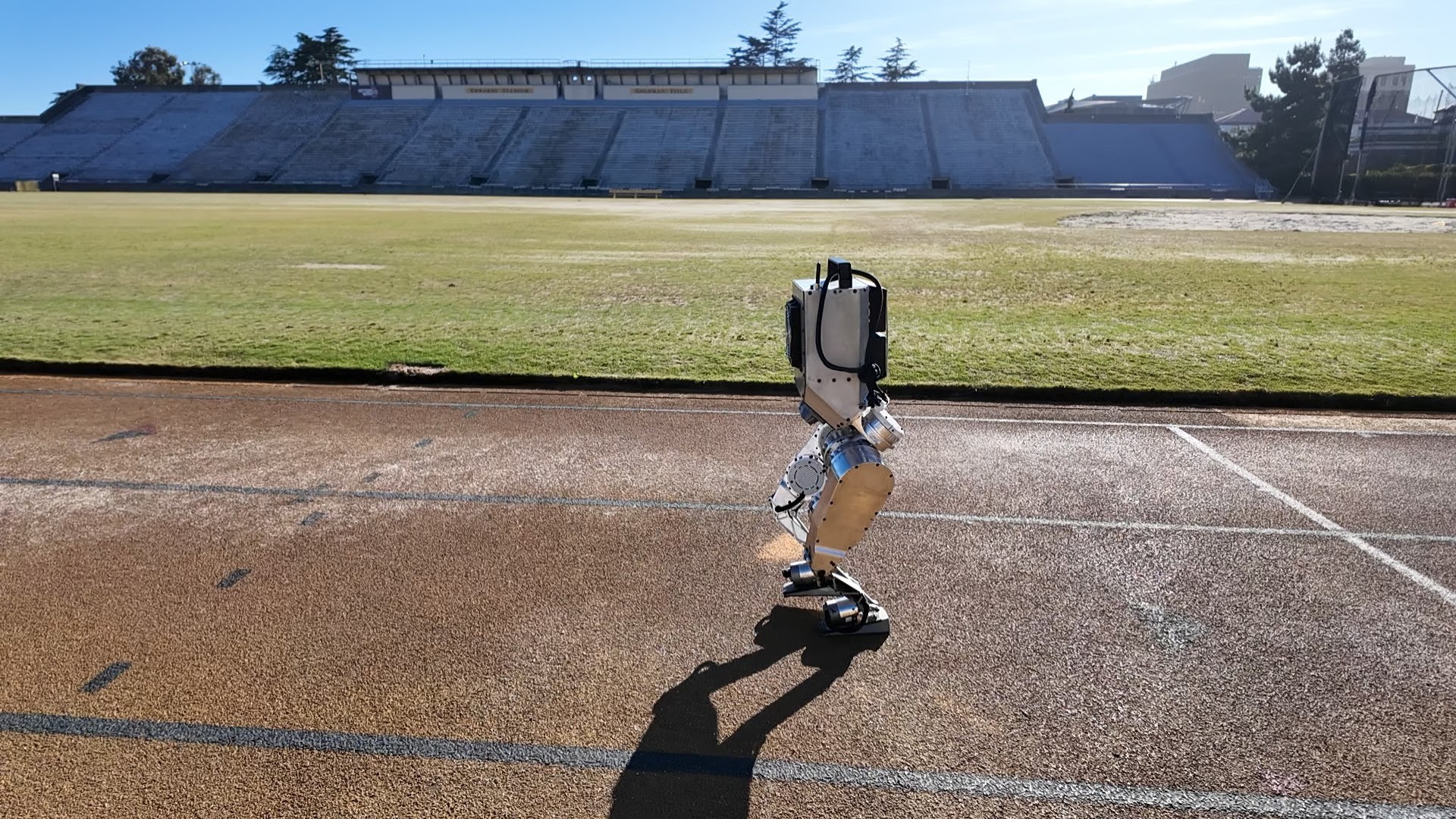}};
    \node[below=0pt of road] (tile) {\includegraphics[width=0.25\width,trim={300 0 300 0},clip]{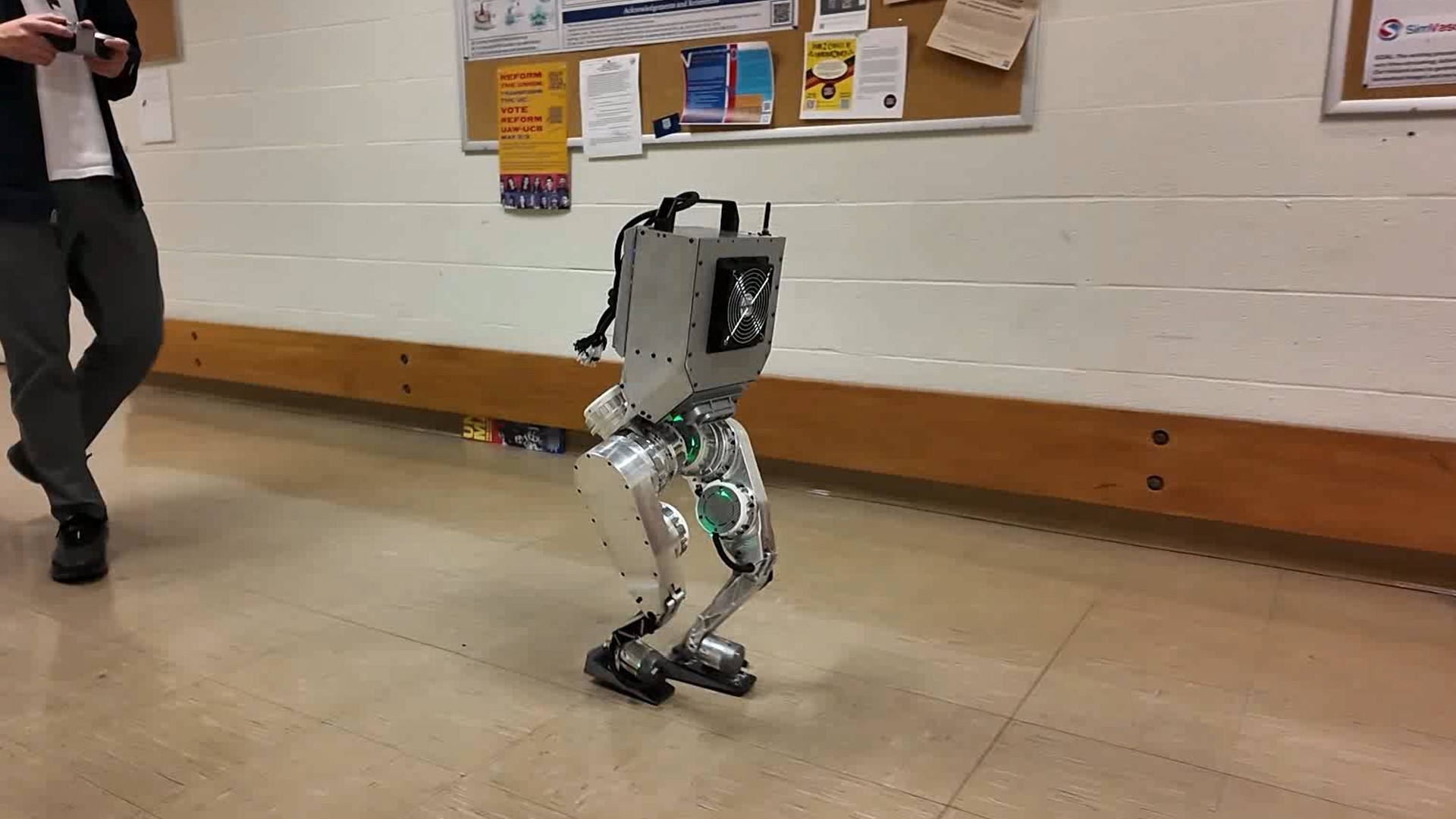}};

    \node[below=\smallgap of bridge.south west, anchor=north west] (uphill) {\includegraphics[width=\width]{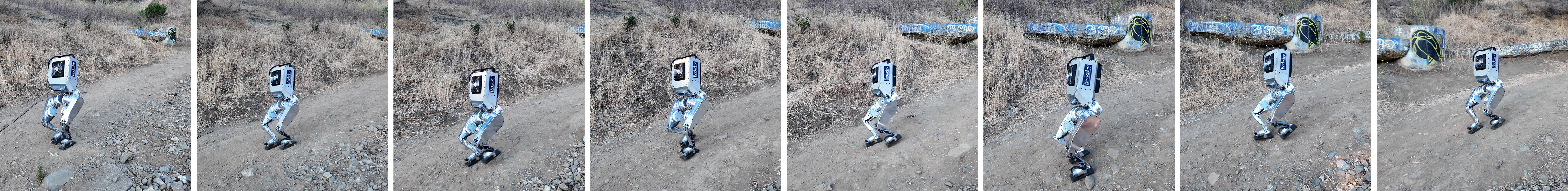}};
    \node[below=\smallgap of uphill] (pathway) {\includegraphics[width=\width]{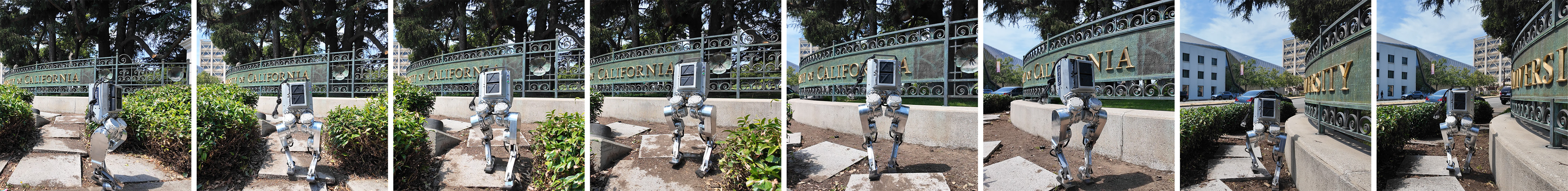}};
    \node[below=\smallgap of pathway] (stairs) {\includegraphics[width=\width]{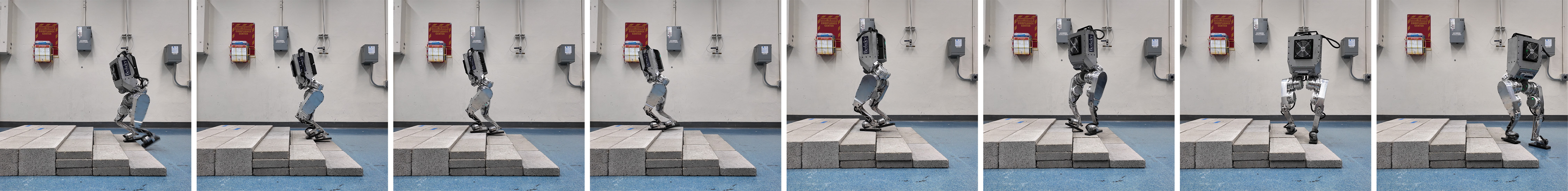}};

    \draw[my_blue] (grass.north west) rectangle (tile.south east);
    \draw[my_yellow] (uphill.south west) rectangle (uphill.north east);
    \draw[my_purple] (pathway.south west) rectangle (pathway.north east);
    \draw[my_green] (stairs.south west) rectangle (stairs.north east);

    \node[white, rectangle, fill=my_blue, inner sep=1pt, anchor=south west, align=center] at (bridge.south west) {\footnotesize (a)};
    \node[white, rectangle, fill=my_yellow, inner sep=1pt, anchor=south west, align=center] at (uphill.south west) {\footnotesize (b)};
    \node[white, rectangle, fill=my_purple, inner sep=1pt, anchor=south west, align=center] at (pathway.south west) {\footnotesize (c)};
    \node[white, rectangle, fill=my_green, inner sep=1pt, anchor=south west, align=center] at (stairs.south west) {\footnotesize (d)};

    \end{tikzpicture}
    \caption{Walking on Various Terrains. (a) The robot walks on eight different types of terrain. (b) The robot climbs a relatively steep and narrow unpaved trail covered with dust and rocks. (c) The robot walks on an uneven pathway. (d) The robot makes a turn on rocky stairs.}
    \label{fig:terrain}
\end{figure}
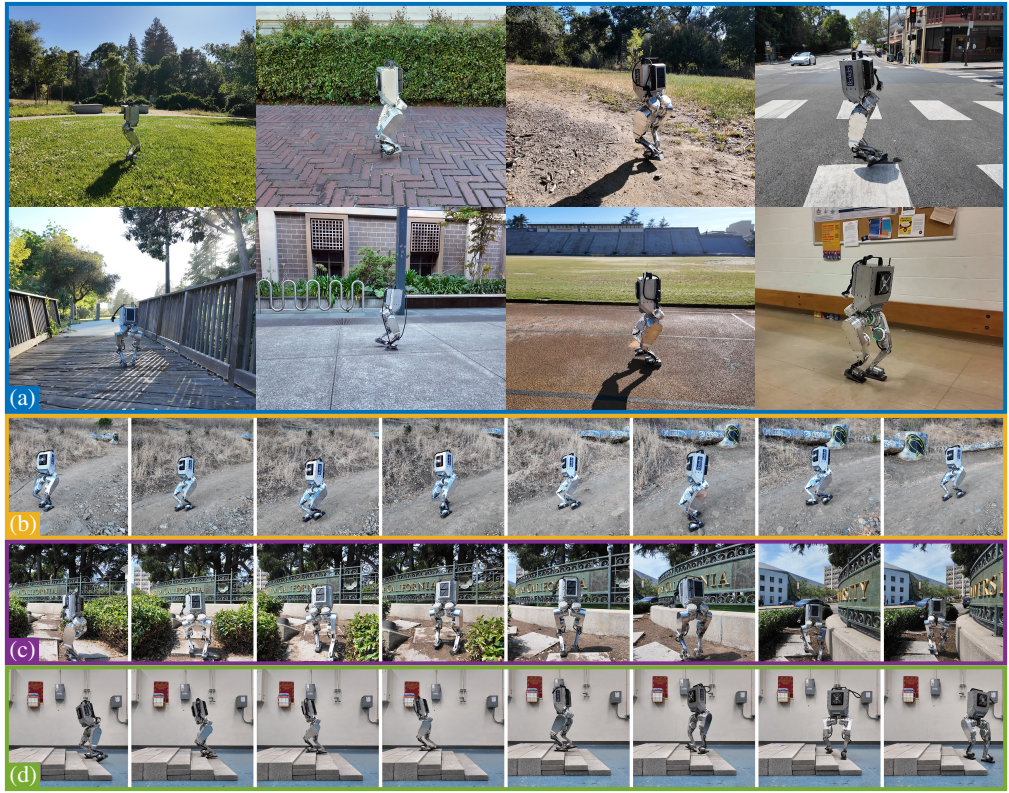

\begin{figure}[t]
    \centering
    \begin{tikzpicture}[line width=1.5pt,inner sep=0pt]
    \let\width\relax
    \newlength{\width}
    \setlength{\width}{0.95\linewidth}
    \def\smallgap{1.5pt}
    \node (lab) {\includegraphics[width=0.5\width]{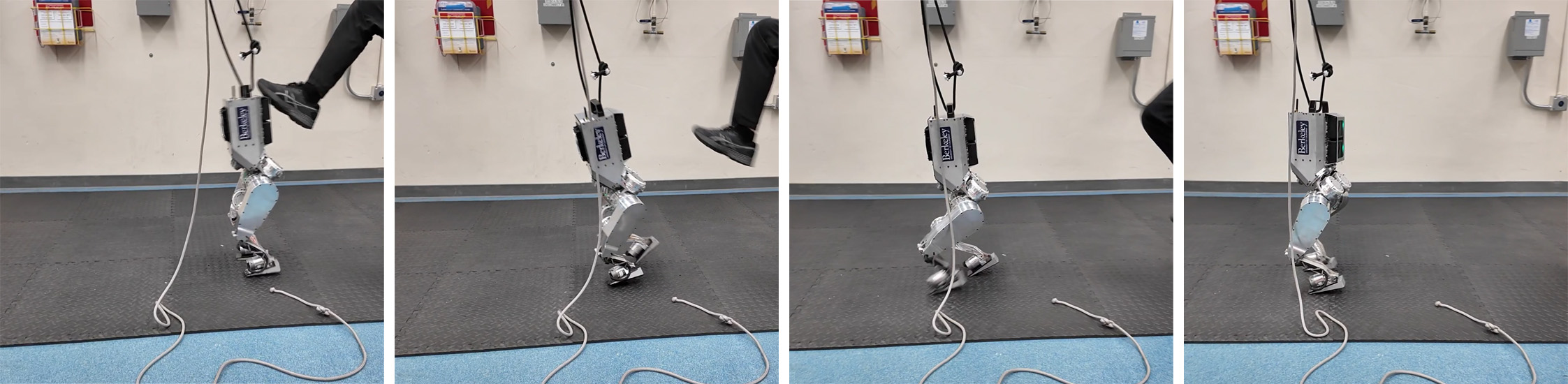}};
    \node[right=\smallgap of lab] (wide) {\includegraphics[width=0.5\width]{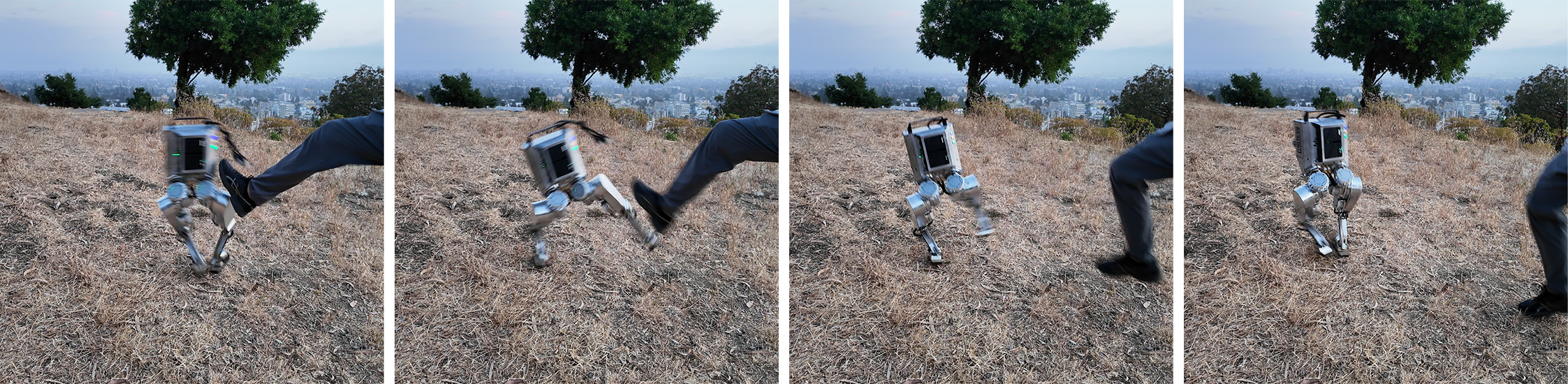}};
    \draw[my_blue] (lab.south west) rectangle (lab.north east);
    \draw[my_yellow] (wide.south west) rectangle (wide.north east);
    \node[white, rectangle, fill=my_blue, inner sep=1pt, anchor=south west, align=center] at (lab.south west) {\footnotesize (a)};
    \node[white, rectangle, fill=my_yellow, inner sep=1pt, anchor=south west, align=center] at (wide.south west) {\footnotesize (b)};
    \end{tikzpicture}
    \caption{Disturbance Rejection. The robot is able to recover from large external perturbations, such as being kicked (a) from behind while walking in the lab, and (b) from the side while walking in the wild.}
    \label{fig:kick}
\end{figure}
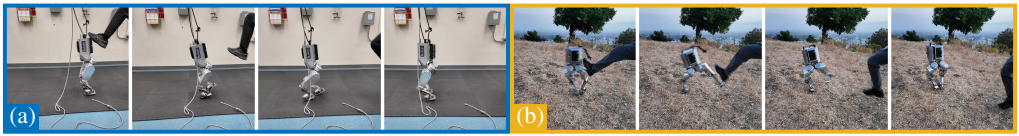

\begin{figure}[t]
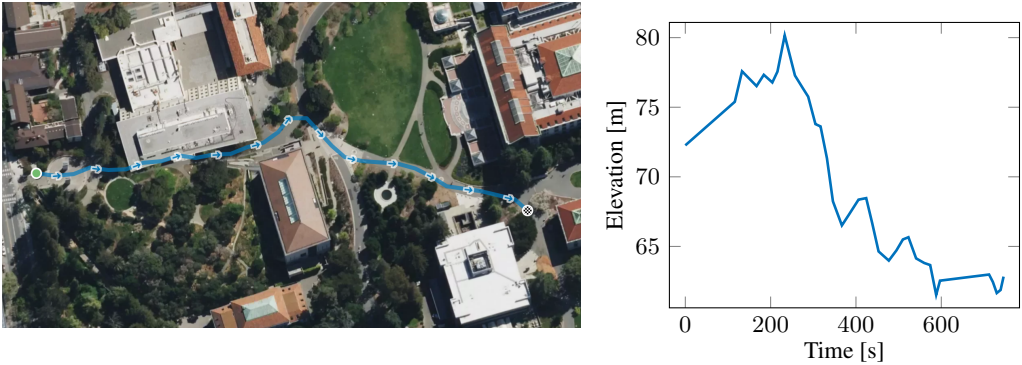

    \centering
    \plotgps
    \caption{Recored GPS visualization of a long distance walking.}
    \label{fig:gps}
\end{figure}

\subsection{Learning Control Performance}
Compared to previous works leveraging advanced architectures, in this work, we emphasize how our minimal design that puts a specific focus on adapting learning-based control algorithms facilitates us to achieve robust and agile locomotion performance with a basic RL controller introduced in Sec.~\ref{sec:control}.

\paragraph{Omnidirectional Walking.}
We train our robot to perform omnidirectional locomotion by following linear velocity commands in sagittal and lateral directions as well as angular velocity commands in yaw. In \figref{fig:omni}, we show examples of walking forward, backward, and turning left and right. In the following paragraphs, we focus on demonstrating the performance of this omnidirectional controller on various terrains and against external perturbations. 

\paragraph{Walking on Various Terrains.}
Perhaps the best demonstration of the advanced performance of a humanoid is its capability to traverse various everyday environments robustly. 
As shown in \figref{fig:terrain}(a), our robot is able to walk robustly on diverse outdoor terrains, such as grass fields, brick sidewalks, unpaved trails, asphalt roads, bridges, concrete roads, running tracks, and tiled surfaces, as well as stairs and inclines. 

Among these environments, we would like to emphasize the two most challenging terrains. 
First, as shown in \figref{fig:terrain}(b) and the accompanying video, we are surprised to find that our robot is able to climb a relatively steep and narrow unpaved trail covered with dust and rocks. This trail is a bit steep to climb even for adults, let alone our robot which resembles only a 5-year-old child in size. Specifically, the incline of the trail is on average 20 degrees, higher than the upward pitch range of the ankle so that it has to go backward to be able to step firmly on the ground with the torso in the upright position. Despite this, our robot is able to walk stably, make turns, and recover from stepping on loose rocks. 

Second, as shown in \figref{fig:terrain}(c), we often find uneven pathways with noticeable gaps and changes in height between the slabs in urban environments. These gaps and slippery slabs require extra attention from children and aged individuals and sometimes cause them to fall over. On this challenging terrain, our robot is able to navigate both forward and backward inside the small pathway across changes in stair heights and recover from slipping. 

In order to further demonstrate uneven terrain, we create a set of rocky stairs with step heights of 4 cm (10\% of full leg length) and find that our robot is able to traverse the stairs smoothly and make turns on them, as seen in \figref{fig:terrain}(d). Being able to handle these challenging terrains shows an advanced performance on locomotion control for our humanoid, even with such a basic RL controller, attributed to the careful adaptations for learning-based control algorithms in the hardware design.

\paragraph{Disturbance Rejection.}
A crucial test of the robustness of the policy and the reliability of the hardware is the ability to recover from external perturbations. We exert instantaneous force randomly by kicking different parts of our robot while it is stepping in place. As shown in \figref{fig:kick}, this perturbation causes a significant deviation from the nominal walking pose, making the robot almost fall over. Nevertheless, our robot is able to respond immediately, regain its stability from the perturbation within a few steps, and resume stepping. 

In addition to the flat ground in the controlled lab environment, we repeat this test in outdoor environments, such as on uneven grass terrains. In these conditions, our robot is also able to recover from heavy external forces, as shown in \figref{fig:kick}(b). This further showcases the robustness of our humanoid robot in real-world scenarios. 

\paragraph{Long Distance Walking}
With the ability to traverse terrains and reject perturbations, the robot is able to perform relatively long-distance walking for several hundred meters over multiple terrains. As shown in \figref{fig:gps}, the robot rambles freely on the campus of UC Berkeley for 10 minutes, traversing a total distance of 364 m with uphills and downhills. 
Furthermore, the robot is able to climb steadily along the rough terrain shown in \figref{fig:terrain}(b) for more than 5 minutes non-stop, covering 96 m in distance and an elevation gain of 10.5 m.
The video of the campus walking can be seen at \url{https://youtu.be/STbB12-oc_w} and the video of walking on rough terrain is at \url{https://youtu.be/Z2Bzslmu7DA}. 

\begin{figure}[t]
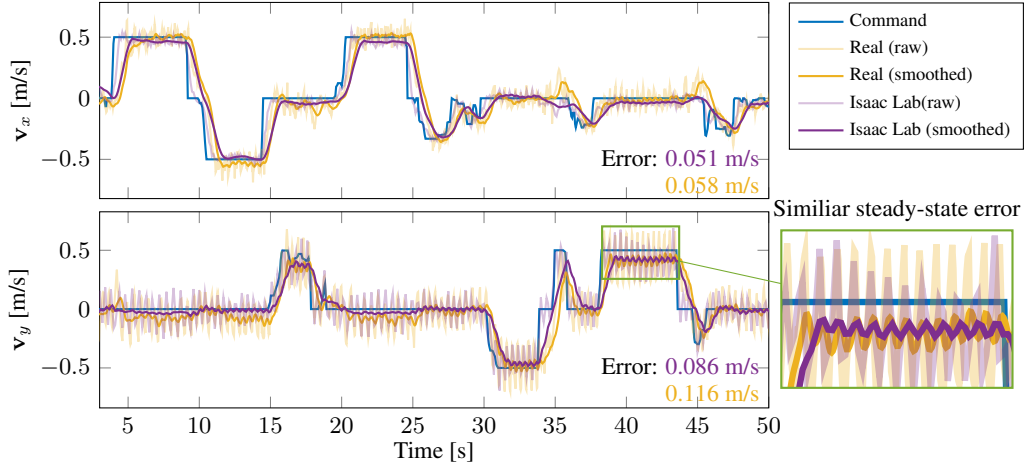

    \centering
    \plottracking
    \caption{Sim-to-real gap evaluation. We show trajectories for commanded (blue) and actual (yellow) base linear velocity. The actual value is smoothed by a moving average filter to better illustrate the steady-state error.}
    \label{fig:tracking}
\end{figure}

\begin{figure}[t]
    \centering
    \begin{tikzpicture}[line width=1.5pt,inner sep=0pt]
    \let\width\relax
    \newlength{\width}
    \setlength{\width}{0.95\linewidth}
    \def\smallgap{1.5pt}
    \node (double) {\includegraphics[width=\width]{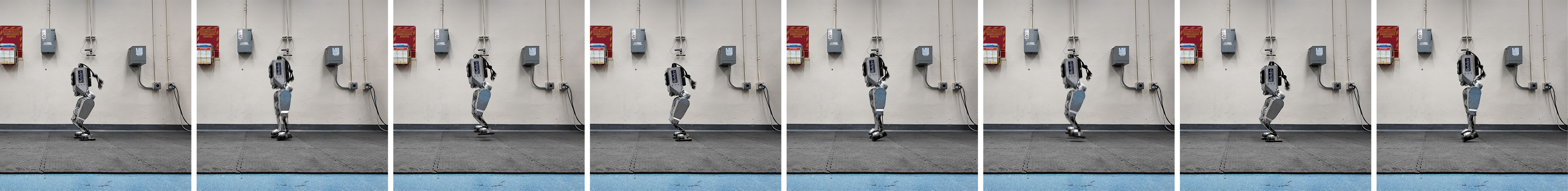}};
    \node[below=\smallgap of double] (single) {\includegraphics[width=\width]{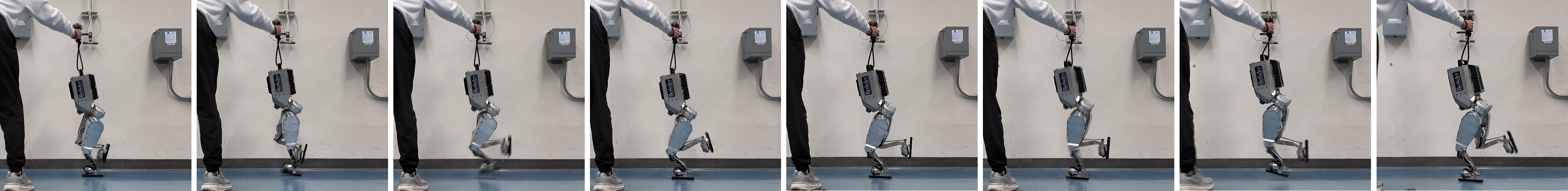}};
    
    \draw[my_blue] (double.south west) rectangle (double.north east);
    \draw[my_yellow] (single.south west) rectangle (single.north east);
    
    \draw[my_purple] ($(double.south west)!2*0.125!(double.south east)+(0,\smallgap)$) rectangle ($(double.north west)!3*0.125!(double.north east)-(0,\smallgap)$);
    \draw[my_purple] ($(double.south west)!5*0.125!(double.south east)+(0,\smallgap)$) rectangle ($(double.north west)!6*0.125!(double.north east)-(0,\smallgap)$);
    \draw[my_purple] ($(single.south west)!2*0.125!(single.south east)+(0,\smallgap)$) rectangle ($(single.north west)!3*0.125!(single.north east)-(0,\smallgap)$);
    \draw[my_purple] ($(single.south west)!6*0.125!(single.south east)+(0,\smallgap)$) rectangle ($(single.north west)!7*0.125!(single.north east)-(0,\smallgap)$);

    \node[white, rectangle, fill=my_blue, inner sep=1pt, anchor=south west, align=center] at (double.south west) {\footnotesize (a)};
    \node[white, rectangle, fill=my_yellow, inner sep=1pt, anchor=south west, align=center] at (single.south west) {\footnotesize (b)};
    \end{tikzpicture}
    \caption{Hopping with (a) both legs and (b) a single leg, with noticeable flight phases. Being able to accomplish dynamic tasks with a simple RL controller shows the small sim-to-real gap of the hardware design. The purple frames indicate that the robot is in the flight phase.}
    \label{fig:hopping}
\end{figure}
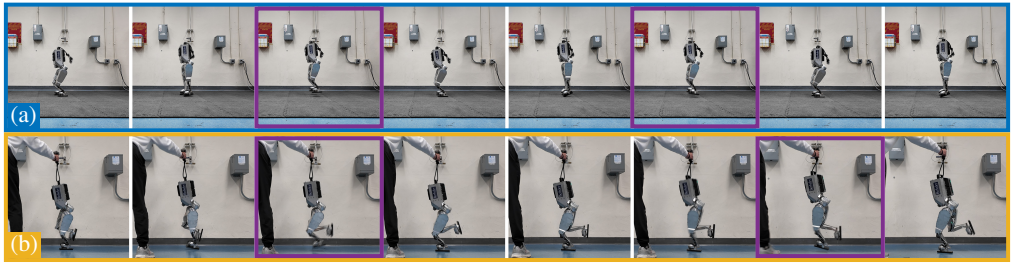

\subsection{Evaluation of Sim-to-Real Transfer}
Because the majority of learning-based algorithms are trained entirely in simulation, the sim-to-real gap becomes a critical component of the performance of learning-based controllers in the real world. We demonstrate the small sim-to-real gap of our robot in two aspects: 
(i) A quantitative analysis of the locomotion task metrics. 
(ii) The ability to perform highly dynamic locomotion tasks. 

First, we present a quantitative analysis of the sim-to-real transfer by plotting the tracking performance with random velocity commands given by the operator. As shown in \figref{fig:tracking}, our robot is able to follow the rapidly changing command closely in both lateral and sagittal directions with small steady-state errors. Over a 60-second trial, the average tracking error in the sagittal direction is 0.051 m/s in simulation and 0.058 m/s on hardware. In the lateral direction, the error is 0.086 m/s in simulation and 0.1156 m/s on hardware, respectively. Note that our RL controller is unable to perform online system identification or adaptation as it does not have access to the history during either training or deployment. Thus, these small differences in tracking errors indicate that the gap between the simulation MDPs during training and the MDPs of the real-world deployment is indeed small, which confirms the narrow sim-to-real gap for our hardware design. 

Second, we showcase the ability to perform highly dynamic motions by demonstrating a hopping controller trained with the same settings as in Sec.~\ref{sec:control} except for the rewards. As shown in \figref{fig:hopping}(a), our robot can perform omnidirectional hops, accelerate, and decelerate while maintaining balance. Notably, the robot further demonstrates exceptional agility by being able to perform hops using only one leg in \figref{fig:hopping}(b), a highly challenging feat. 
Although a safety rope is used and minor balance assistance is needed during single-leg hopping experiments, the rope is mostly slack, and the robot is able to maintain its balance on its own.
Compared to complex algorithm designs in prior works, this further shows that the hardware design facilitates us to perform agile motions with simple algorithmic design. 

\subsection{Hardware Reliability}
Lastly, hardware reliability against ground impacts is vital for learning-based approaches. Throughout this work, we recorded a total of 38 times of our robot falling over on various terrains including concrete pavements and unpaved roads, shown in \tabref{tab:falls_number} in the Appendix. 
Thanks to the reliable and lightweight design, we did not experience any damage to the hardware itself except for two failures caused by loose screws and glue. In most fallovers, we are able to reset the robot and resume the control policy within 3 to 5 seconds. The ability to reset easily and rapidly not only relieves the burden of experiments but more importantly, is necessary for the ultimate goal of scalable real-world deployment. 

\section{Limitations}
\label{sec:limitation}

Major limitations of this work include the omission of arms for simplicity since the main research topic of mid-scale humanoids still focuses on locomotion tasks. The range of motion, backlash, weight, and mechanical strength, will be further improved after a few hardware iterations. To further minimize the sim-to-real gap for more dynamic motion, detailed system identification for torque-current non-linear mapping near saturated torque should be performed. Motor region of work \cite{shin2023actuator}, and heat protection should be simulated during training. In the future, the platform will be equipped with two 4 or 6 DoFs arms and enough power to perform dynamic tasks such as backflipping. 

\section{Conclusion}
This work introduced the \robotname, a reliable and low-cost research platform for learning-based bipedal locomotion control with a narrow sim-to-real gap. Our in-house-built humanoid robot specializes in accommodating learning-based control algorithms, featuring low simulation complexity, anthropomorphic ranges of motions, and high reliability against falls and impacts. 
Designed with lightweight materials, it greatly reduces the burden of conducting hardware experiments.
Being able to perform robust outdoor experiments over various terrains and ground conditions with only a minimally designed RL algorithm further underscores the efficacy of our platform for learning-based control and its small sim-to-real gap.
Our policy, without history or phase signal as input, is able to withstand large, random external perturbations and perform omnidirectional locomotion over challenging terrains. 
Notably, it demonstrates the ability to walk long distances on campus, climb steadily along steep and narrow unpaved trails, and hop with a single leg, a highly dynamic feat.
As a reliable, low-cost research platform, the ultimate goal is to deploy scalably for learning in the real world.



\acknowledgments{
This work was supported in part by The AI Institute. We would like to thank Jiaze Cai for the generous help with the experiments, and Yufeng Chi for suggesting the name of the robot. We'd also like to express our gratitude to Prof. Wei Zhang and Pan Motor for their valuable discussions and assistance with the actuators.}


\bibliography{reference}  

\newpage

\appendix
\section*{Appendix}

\section{Reward Function}
In this section, we provide the detailed reward functions used to train our policy. 

\subsection{Walking}
The reward function design for walking has four parts. The first part includes tracking terms, implemented as the $L^2$ norm of the difference between the desired and actual linear velocities in the sagittal and lateral directions, as well as the angular velocity in the yaw. 

The second part is the smoothing terms where we penalize non-zero values in the linear velocity in the vertical direction and angular velocities in both roll and pitch. The joint torques and action rates are also penalized. These terms help improve the smoothness of the policy. 

Furthermore, we regularize the hip and knee joints with respect to their nominal positions and body orientation with upright orientation. We also set a soft limit for the actuators, over which the actions will be penalized. These regularization terms are beneficial in preventing aggressive and dangerous motions the policy might learn. 

Lastly, we include gait quality terms necessary for exhibiting reasonable walking gaits. These terms encourage feet to stay longer in the air \citep{rudin2022learning}, to not slip on the ground \citep{mittal2023orbit}, and to keep contact forces under a threshold to protect the gearboxes and other hardware.


\subsection{Hopping}
The reward function for hopping is slightly modified from the walking task. First, instead of penalizing vertical linear velocity, we encourage positive linear velocity in vertical direction using a ReLU function, namely, $r_{vz} = \text{ReLU}(\mathbf{v}_z)$. Second, we do not limit knee joints and hip joints in pitch as they are necessary in providing a large upward acceleration in hopping. Additionally, in single-leg hopping, we penalize the in-air leg contacting with the ground. Apart from these, the other terms stay the same as the walking task. 

\section{Outdoor Failure Counts Throughout the Project}
Throughout the entire project, we record the failure counts over different terrains as proof of the durability of the robotic hardware. Note that this represents failures during testing and debugging of the hardware, but not the experiments presented above. 

\begin{table}[ht]
\centering
\caption{Number of \textbf{Recorded} Falls on Different Surfaces.}
\label{tab:falls_number}
\begin{tabular}{c|cccc}
\toprule
Surface & Stone Brick Road & Grassland & Running Track & Unpaved Road \\
\midrule
Number  & 6                & 14        & 3             & 15           \\ 
\bottomrule
\end{tabular}
\end{table}

\section{Joint Ranges of the Hardware}
As discussed in Sec. \ref{sec:anthropomorphic}, our hardware follows an anthropomorphic design to approach the range of human movements as much as possible. The ranges for each of the 6 DoFs are recorded in the table below, 
\begin{table}[ht]
\centering
\caption{Comparsion Between Ranges of Motion Humand Joint and Proposed Robot (right leg). Data from~\citep{houglum2011brunnstrom}, \citep{hamilton2011kinesiology}, \citep{ball1996technique}.}
\label{tab:human_joint_angle}

\begin{threeparttable}
\begin{tabular}{c|cccccc}
\toprule
Joint Names               & HR          & HAA         & HFE         & KFE       & FFE        & FAA        \\
\midrule
Human [\textdegree]       & $[-50, 40]$ & $[-40, 20]$ & $[-110,30]$ & $[0,150]$ & $[-20,50]$ & $[-30,18]$ \\
Proposed [\textdegree]    & $[-35, 35]$ & $[-35, 35]$ & $[-100,30]$ & $[0,120]$ & $[-30,70]$ & $[-30,30]$ \\
Coverage Rate             & $77.8\%$    & $91.6\%$    & $92.9\%$    & $80.0\%$  & $100.0\%$  & $100.0\%$ \\
\bottomrule
\end{tabular}

\end{threeparttable}
\end{table}
The joint names and their definitions are as follows:
\begin{itemize}
    \item HR: Hip Rotation
    \item HAA: Hip Abduction/Adduction
    \item HFE: Hip Flexion/Extension
    \item KFE: Knee Flexion/Extension
    \item FFE: Foot Flexion/Extension
    \item FAA: Foot Abduction/Adduction
\end{itemize}

\begin{figure}[ht]
    \centering
    \includegraphics[width=0.8\linewidth,page=4, trim={0 0 0 0}, clip]{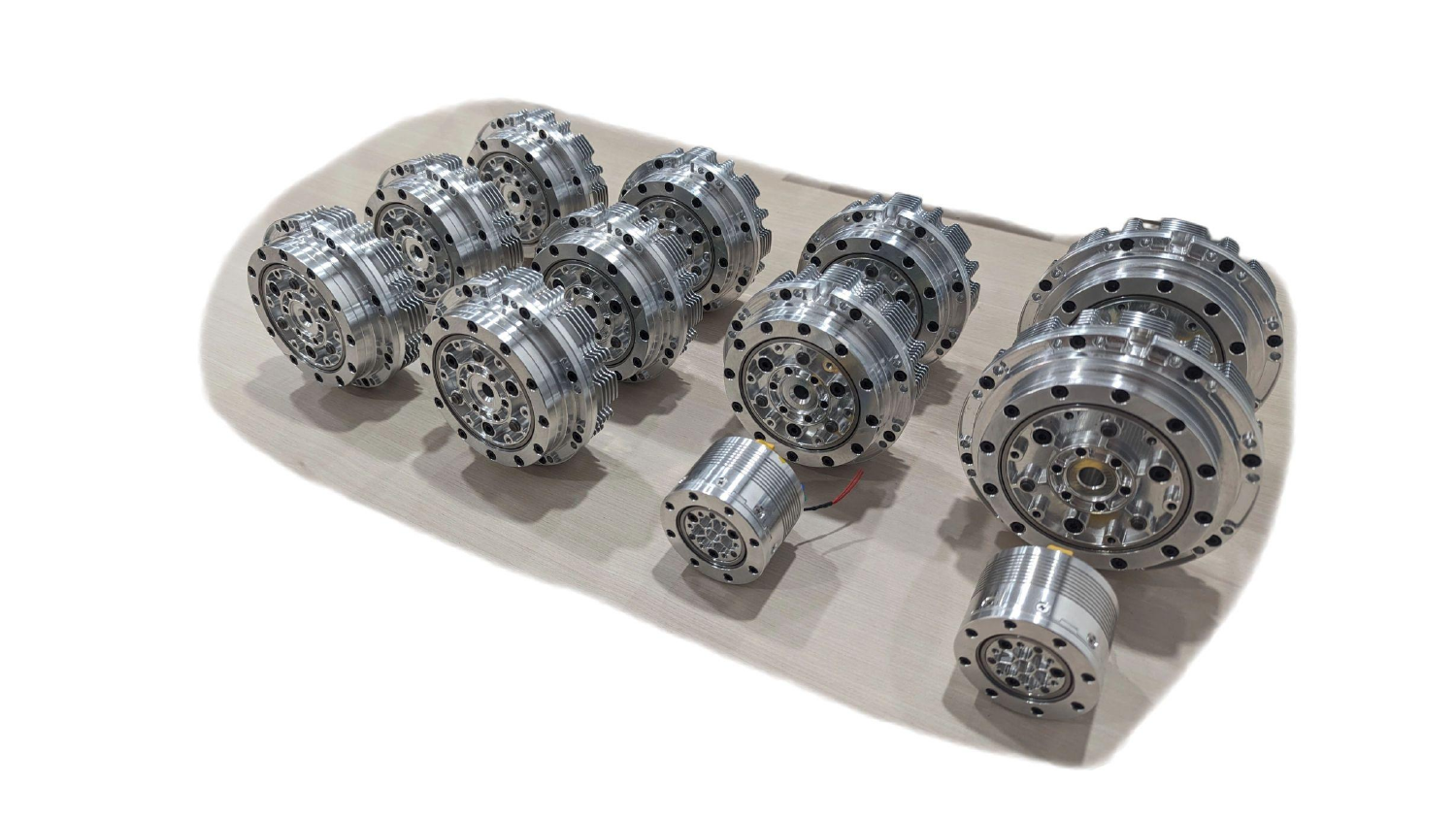}
    \caption{Totally 12 Actuators Used in the Robot.}
    \label{fig:actuators_in_robot}
\end{figure}

\section{ Dynamics Randomization Details}
As discussed in Sec. \ref{sec:sim2real}, we designed dynamics randomization carefully to best fit the actual hardware. The ranges are summarized in Table \ref{tab:domain_rand} below, 

\begin{table}[ht]
\centering
\resizebox{0.99\columnwidth}{!}{
\begin{tabular}{c|ccccccc}
\toprule
Dynamics Terms & Friction & Restitution & Base Mass & Linkage Mass & Joint Friction & Joint Armature & Default Joint Pos\\
\midrule
Low  & 0.2  & 0.0 & -1.0 & x0.9 & x0.9 & x1.0 & -0.05\\
High & 1.25 & 0.1 & +1.0 & x1.1 & x1.1 & x1.05 & 0.05\\
\bottomrule
\end{tabular}
}
\vspace{2mm}
\resizebox{0.9\columnwidth}{!}{
\begin{tabular}{c|cccccccc}
\toprule
Noise Terms   & Lin Vel & Ang Vel & IMU  & Hip Joints Pos & KFE Pos & FFE Pos & FAA Pos & Joints Vel\\
\midrule
Range ($\pm$) & 0.1     & 0.2     & 0.05 & 0.03           & 0.05    & 0.08    & 0.03    & 1.5        \\
\bottomrule
\end{tabular}
}
\vspace{2mm}
\caption{List of domain randomizations. After system identification on the in-house designed hardware, we provide a small range of 6 dynamics parameters and 8 noise terms that minimize the sim-to-real gap.}
\label{tab:domain_rand}
\end{table}

\end{document}